\documentclass{article} 
\usepackage[dvipsnames]{xcolor}
\usepackage{iclr2025_conference,times}

\usepackage{amsmath,amsfonts,bm}

\def\eqref#1{equation~\ref{#1}}

\def\1{\bm{1}}

\DeclareMathAlphabet{\mathsfit}{\encodingdefault}{\sfdefault}{m}{sl}
\SetMathAlphabet{\mathsfit}{bold}{\encodingdefault}{\sfdefault}{bx}{n}

\usepackage{hyperref}
\usepackage{url}
\usepackage{xspace}
\usepackage{enumitem}

\usepackage{booktabs}
\usepackage{multirow}
\usepackage{graphicx}
\usepackage{arydshln}
\usepackage{colortbl}
\usepackage{algorithm}
\usepackage{wrapfig}
\usepackage{algpseudocode}

\definecolor{backgreen}{RGB}{226, 240, 217}

\title{SPaR: Self-Play with Tree-Search Refinement to Improve Instruction-Following in Large Language Models}

\author{Jiale Cheng$^{1,2}$\thanks{\ Equal contributions.}\ \thanks{Work done when JC and YL interned at Zhipu AI.} , Xiao Liu$^{2,3}$\footnotemark[1] , Cunxiang Wang$^{2,3}$ , Xiaotao Gu$^{2}$ , Yida Lu$^{1,2}$\footnotemark[2] , Dan Zhang$^{3}$ , \\ \textbf{Yuxiao Dong$^{3}$ , Jie Tang$^{3}$ , Hongning Wang$^1$ , Minlie Huang$^1$\thanks{Corresponding author}}
\\
$^1$The Conversational Artificial Intelligence (CoAI) Group, Tsinghua University \\$^2$Zhipu AI\\
$^3$The Knowledge Engineering Group (KEG), Tsinghua University\\
\small{\texttt{{ chengjl23@mails.tsinghua.edu.cn,}}} \small{\texttt{aihuang@tsinghua.edu.cn}}\\
}

\newcommand{\hhide}[1]{}
\newcommand{\model}[0]{\textsc{SPaR}\xspace}

\newcommand{\hnote}[1]{{\color{red}{[WHN: #1]}}}

\iclrfinalcopy 
\begin{document}

\maketitle

\begin{abstract}
    
Instruction-following is a fundamental capability of language models, requiring the model to recognize even the most subtle requirements in the instructions and accurately reflect them in its output.
Such an ability is well-suited for and often optimized by preference learning.
However, existing methods often directly sample multiple independent responses from the model when creating preference pairs.
Such practice can introduce content variations irrelevant to whether the instruction is precisely followed (e.g., different expressions about the same semantic), interfering with the goal of teaching models to recognize the key differences that lead to improved instruction following.
In light of this, we introduce \model, a self-play framework integrating tree-search self-refinement to yield valid and comparable preference pairs free from distractions.
By playing against itself, an LLM employs a tree-search strategy to refine its previous responses with respect to the instruction while minimizing unnecessary variations.
Our experiments show that a LLaMA3-8B model, trained over three iterations guided by \model, surpasses GPT-4-Turbo on the IFEval benchmark without losing general capabilities. 
Furthermore, \model demonstrates promising scalability, greatly enhancing models like GLM-4-9B and LLaMA3-70B.
We also identify how inference scaling in tree search would impact model performance.
Our code and data are publicly available at \url{https://github.com/thu-coai/SPaR}.

\end{abstract}

\begin{figure}[htbp]
    \centering
\includegraphics[width=0.95\linewidth]{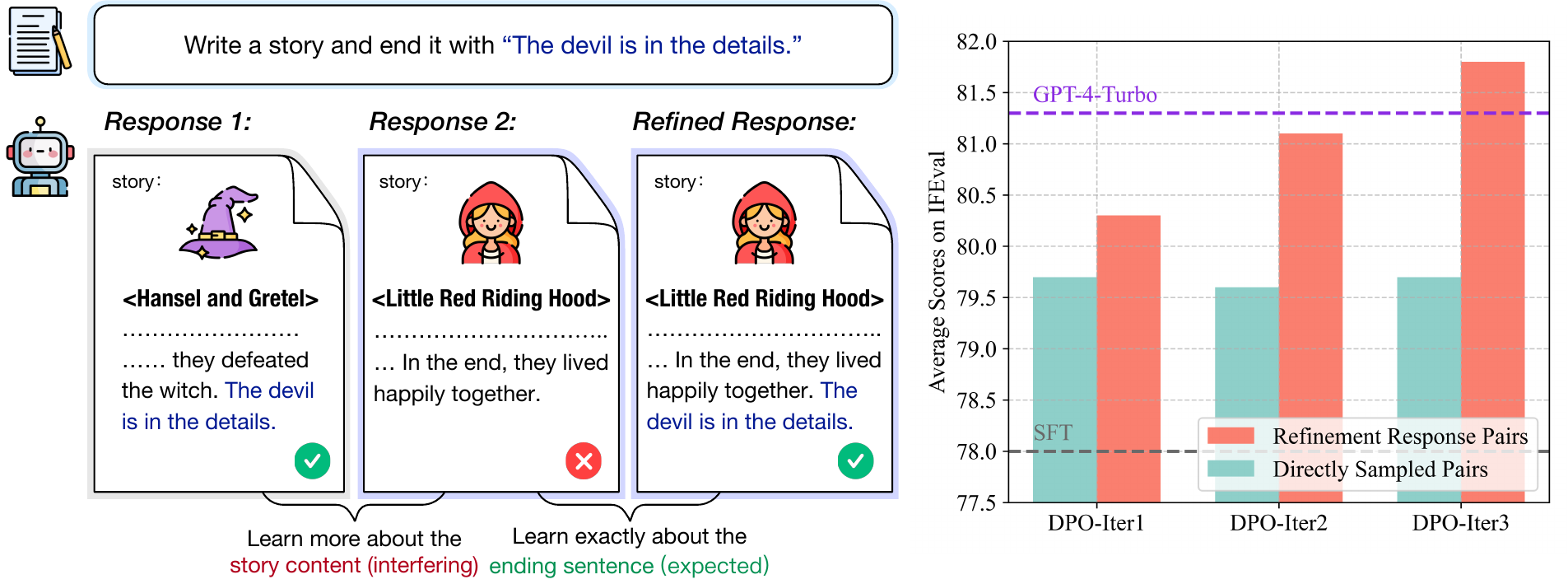}
    \caption{An example of the interfering factors (\textit{story content}) in independently sampled multiple responses (Left). Refined response pairs exclude these factors, highlight the key difference (\textit{ending sentence}), and lead to improved performance on iteratively trained LLaMA3-8B-Instruct (Right).}
    \label{fig: abs}
    \vspace{-3mm}
\end{figure}

\section{Introduction}

To date, Large Language Models (LLMs) have achieved great success in a wide range of tasks \citep{brown2020language, zeng2022glm, chowdhery2023palm, touvron2023llama, glm2024chatglm}.
As LLMs are applied to various scenarios, their instruction-following capability becomes crucial \citep{ouyang2022training, bai2022training}, especially to follow instructions with multiple constraints \citep{zeng2023evaluating, zhou2023instruction, jiang2023followbench}. The failure to accurately follow instructions can even lead to safety issues \citep{ruan2023identifying}.

Subtle nuances can determine the success of instruction-following tasks \citep{zhou2023instruction}, making preference learning \citep{rafailov2024direct, hou2024chatglm} a well-suited solution.
However, existing methods usually sample multiple independent responses from the target model \citep{yuan2024self, wu2024meta, dong2024self}, inadvertently introducing irrelevant variations to whether the instruction was successfully followed.
As illustrated in Figure \ref{fig: abs}, given the instruction: ``Write a story and end it with \textit{The devil is in the details}", 
sampling multiple independent responses from an LLM can result in responses as different as the story \textit{Little Red Riding Hood} vs. \textit{Hansel and Gretel}. This variation in the narrative content can interfere with the model's ability to learn how to realize the critical requirement—the specified ending sentence—and ultimately mislead the comparison within the preference pair. Therefore, effective learning from preference pairs necessitates excluding these extraneous factors and focusing on the key differences that drive the success of instruction-following.

In this paper, we propose \model, a self-play method integrated with tree-search refinement to enhance instruction-following capabilities of LLMs.
The key lies in iteratively teaching LLMs to learn instruction-following from nuances by playing against itself with structured tree search.
In each turn of self-play, an LLM takes two different roles: the actor and the refiner, which are both initialized from the same model.
The actor executes complex instructions while the refiner critiques and refines the actor's responses.
During the iteration, we first collect the actor's responses which fail to follow the instructions accurately, as judged by the refiner.
Starting from those failed responses, we apply a tree-search algorithm for refinement, which ensures consistent improvements against previous turns and naturally creates valid comparison counterparts for model training.

We conduct experiments on several LLMs, LLaMA3 series \citep{llama3}, GLM-4-9B \citep{glm2024chatglm}, and Mistral-7B-Instruct \citep{jiang2023mistral}, over multiple iterations.
Through extensive experiments, we demonstrate significant improvements in the models' instruction-following capability, outperforming other self-improvement methods such as self-rewarding \citep{yuan2024self} and meta-rewarding \citep{wu2024meta}. Notably, after three iterations, \model improves LLaMA3-8B-Instruct over GPT-4-Turbo on the IFEval benchmark \citep{zhou2023instruction}. 
Moreover, scaling test-time compute by integrating tree-search refinement during inference can further improve the quality of instruction following.
Additionally, we find that with several iterations, the refiner's judgment and refinement capabilities can match or even exceed those of the distilled LLM, indicating great potential for continuous self-improvement without being limited by the initial bootstrapping data. Ablation studies demonstrate the importance of each component within our framework. Importantly, our method does not degrade performance on general benchmarks.
In summary, our contributions are:
\begin{itemize}[leftmargin=1.5em,itemsep=0pt,parsep=0.2em,topsep=0.1em,partopsep=0.0em]
    \item We reveal that preference pairs derived from independently sampled responses often contain interfering factors, hampering preference learning to improve instruction following. 
    As a result, a performing solution has to minimize such interference and highlight the key differences contributing to the success of instruction following.
    \item We introduce \model, a novel self-play framework that enables continuous self-improvement in instruction-following tasks. Through three iterations, our method boosts LLaMA3-8B-Instruct to achieve GPT4-level performance and scales effectively to enhance LLaMA3-70B-Instruct.
    \item We construct a high-quality dataset with 43K complex instruction-following prompts and an SFT dataset that can improve the instruction-following capabilities of LLMs.    
\end{itemize}


\section{Method}

\begin{figure}[htbp]
    \centering
    \includegraphics[width=0.9\linewidth]{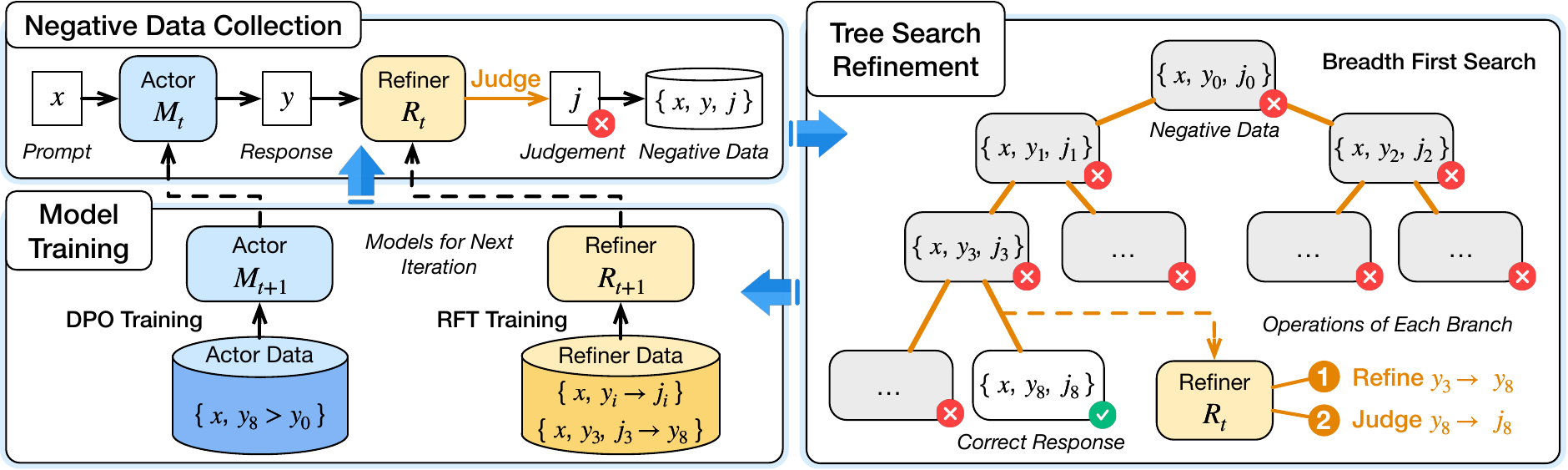}
    \caption{\model iterative training framework. At iteration $t$, the refiner $R_t$ first judges the generated responses from the actor $M_t$ to collect negative data. Next, a tree-search algorithm is employed to refine these imperfect responses. Finally, using the data from the above steps, we can optimize the actor and refiner for the next iteration, aiming for continuous self-improvement.}
    \label{fig: method}
    \vspace{-5mm}
\end{figure}


We introduce \model, an automated and scalable approach designed for self-improvement of instruction-following tasks through self-play. The core idea is to create paired responses with minimal irrelevant variations, thereby highlighting the key differences that manifest the success of instruction-following.

\subsection{Overall Framework}

The overall framework of \model is illustrated in Figure \ref{fig: method}. 
Briefly, our framework involves an actor model and a refiner model, which are both initialized from the same base model. 
The actor generates responses to given instructions while the refiner judges and refines these responses. This iterative self-play process, involving response generation, judgment, and refinement, fosters continuous self-improvement.

Formally, in each iteration, given an instruction $x$ from the prompt set, the actor generates a response $y$. The refiner identifies the responses that do not follow the instructions accurately, termed as negative responses. Our objective is to refine the negative response (represented as $y_0$ in Figure \ref{fig: method}) into a correct response (represented as $y_8$ in the figure). These generated refinement pairs, e.g., $(x, y_8 > y_0)$, are collected and used to optimize the actor via Direct Preference Optimization (DPO) \citep{rafailov2024direct}. Simultaneously, we apply Rejection-sampling Fine-Tuning (RFT) \citep{yuan2023scaling} to improve the refiner. This process prepares both models for the next iteration of self-improvement.

In this iterative process, we face two major challenges: the scarcity of complex instruction-following data and the difficulty of achieving successful refinements.
To address the lack of high-quality, multi-constraint instruction-following datasets, we generate complex instructions using a taxonomy-based approach and create corresponding SFT datasets to initialize the actor and refiner models (\S\ref{para: data construction}).
To ensure a high success rate in refining negative responses, we employ a tree search strategy that systematically explores refinement paths and facilitates subsequent training (\S\ref{para: tree search refinement}).

\subsection{Data Construction} \label{para: data construction}

\subsubsection{Prompt Creation} \label{para: prompt creation}
Given the scarcity of high-quality data for instruction-following tasks, especially those with multiple constraints, we start by creating a high-quality dataset of instruction-following prompts.

\vspace{-3mm}
\paragraph{Seed Prompts.}
To ensure the quality and diversity of our dataset, and to prevent issues like insufficient diversity or even model collapse \citep{liu2024best, shumailov2024ai}, we use a seed set of prompts derived from the Infinity-Instruct dataset \citep{zhao2024iidoptimizinginstructionlearning}, which contains ten million high-quality conversations. 
After applying rule-based filtering based on length, keywords, and self-BLEU, we obtain approximately 50k seed prompts.
\vspace{-3mm}
\paragraph{Taxonomy-based Prompt Construction.}  Complex prompts constructed without human intervention tend to be poorly diversified, as the types of constraints added are often distributed unevenly \citep{sun2024conifer}. Therefore, we adopt a taxonomy-based mechanism to make constraint types comprehensive and balanced.
Our taxonomy for instruction-following constraints is derived from \cite{cheng2024autodetect} and further refined to be more comprehensive.
After building the constraint taxonomy, we employ it to construct complex instruction-following tasks based on seed prompts. We sample a main constraint type and employ a strong LLM 
to add several other constraints to make the original prompt more complex. 
Moreover, we leverage the strong LLM to assess the validity of the generated prompt, ensuring that the constraints do not conflict with each other or create unreasonable scenarios with the original task.
The detailed taxonomy and prompt can be found in Appendix \ref{appendix: prompt and taxonomy}.

\subsubsection{Actor and Refiner Initialization} \label{para: model initialization}
The taxonomy-based prompt construction results in about 43k prompts. We utilize 8k prompts for actor initialization, another 5k for the refiner, and save 30k for further self-play training. 

\paragraph{Actor Data Creation.} To bootstrap the actor model with strong instruction-following capabilities, we first collect a strong LLM's responses to these complex prompts, thereby producing supervised fine-tuning (SFT) data $(x, y) \in D_\text{Actor}$ for the actor model, where $x$ is the complex instruction and $y$ is the strong LLM's response. Then, we fine-tune the base model to get an initial actor $M_{0}$.

\paragraph{Refiner Data Creation.} To bootstrap the refiner model with strong judgment and refinement capability, we sample responses from the initial actor $M_0$. Then, we collect the judgments from a strong LLM to form a dataset, $(x, y, j) \in D_\text{JSFT}$.
We collect responses that are judged not to accurately follow instructions and term them as negative responses.
For these negative responses, we use the strong LLM to correct them with minimal revisions to avoid irrelevant variations. 
In this way, we get a refinement dataset, $(x, y_\text{negative}, j, y_\text{refined}) \in D_\text{RSFT}$. The refiner is then trained with $D_\text{Refiner} = D_\text{JSFT} \cup D_\text{RSFT}$ to create the initial refiner $R_0$.

\paragraph{Training Strategy.}
For both actor and refiner models, we use standard supervised fine-tuning with the loss function:
\begin{equation} \label{loss: SFT}
    \mathcal{L}=-\frac{1}{N}\sum_{i=1}^N\text{log}P(r_i|q,r_{<i}), 
\end{equation} 
where $q$ denotes the input, $r$ signifies the target response, and $N$ represents the length of $r$.
For actor training, we have input $q = x$ and target $r = y$. When it comes to the refiner, we use input $q = (x, y)$ and target $r = j$ for $D_\text{JSFT}$, and input $q = (x, y_\text{negative}, j)$ and target $r = y_\text{refined}$ for $D_\text{RSFT}$.

\subsection{Tree-Search Integrated Self-Play Training} \label{para: tree search refinement}
After initializing the actor and refiner models, we embark on an iterative process for continuous self-improvement.
In each iteration, we first collect the negative data, where the responses fail to accurately follow the instructions (\S\ref{para: wrong response collection}). Then, we utilize a tree-search algorithm to refine the negative responses (\S\ref{para: tree search refinemen}) and form the training data for the next iteration of the actor (\S\ref{para: actor dpo}) and refiner (\S\ref{para: reinfer rft}).
This iterative self-play pipeline allows us to continuously improve both models.

\subsubsection{Negative Data Collection} \label{para: wrong response collection}
For each prompt $x$, we first sample $K$ responses $\{y_1, y_2, \ldots, y_K\}$ from the actor model. This step ensures that there are enough negative responses to support subsequent learning. 
Then, for each prompt and response pair, we utilize the refiner to generate a judgment, which contains two parts: a label suggesting whether the response follows the instruction and an explanation about the assessment. To make this judgment more accurate, we incorporate the self-consistency mechanism \citep{wang2022self}, which is also applied in the subsequent refinement process. 
Specifically, we obtain multiple judgments from the refiner and determine the final label through majority voting, as detailed in Appendix \ref{appendix: exp}. 
After majority voting, we randomly select one judgment that matches the voted label to serve as the final judgment.
This process allows us to identify challenging prompts that elicit responses that do not accurately follow the instructions, yielding tuples in the form of $(x, y_\text{negative}, j)$, where $y_\text{negative}$ is the incorrect response and $j$ is its corresponding judgment. 

\subsubsection{Tree-Search Refinement} \label{para: tree search refinemen} 
After collecting these negative instances, the core step is to refine the responses to form preference pairs. 
These self-refined pairs are crucial for highlighting the subtle differences that can determine the success of instruction-following tasks, thereby facilitating effective learning.
Given that direct refinement often results in a low success rate, we employ a tree-search approach. We implement both breadth-first search (BFS) and depth-first search (DFS) strategies for this refinement. Detailed algorithms for these methods are provided in Appendix \ref{appendix: tree search}.

To illustrate our process, we take BFS as an example and illustrate the procedure in Figure \ref{fig: method}. 
Starting with an incorrect instruction-response pair and its judgment as the root node, we expand the search tree level-by-level until a correct response is found.
At each intermediate node, we generate potential refinements for the current response and evaluate its correctness using the refiner.
The number of generated refinements corresponds to the number of branches. Specifically, at a level of the tree, the refiner:
1). generates potential refinements for each node in the current level; 2). judges the correctness of these refinements. This creates a set of child nodes with new responses and their corresponding judgments.
The search process continues until we obtain a tuple $(x, y_\text{negative}, y_\text{refined})$, where $y_\text{refined}$ is the newly refined, correct response. 
Importantly, \model combines the strengths of both tree-search and self-refinement, exploring multiple refinement paths while minimizing the interfering factors, producing effective preference learning data.


\subsubsection{Actor Training} \label{para: actor dpo}
To optimize the actor model, we leverage the refinement pairs for preference learning using DPO.
At iteration $t$, we train the actor model $M_{t}$ with refinement pairs $(y_\text{negative}, y_\text{refined})$, treating $y_\text{negative}$ as the rejected response ($y_l$) and $y_\text{refined}$ as the chosen response ($y_w$). The training dataset is denoted as $D_\text{dpo}^{t}$ and the DPO loss is described as follows:
\begin{equation}
    \mathcal{L}_{\text{DPO}} (\pi_{\theta}^{\text{t}}; \pi_{\text{ref}}) = - \mathbb{E}_{(x, y_w, y_l) \sim D_{\text{dpo}}^{t}} \left[ \log \sigma \left( \beta \log \frac{\pi_{\theta}^{\text{t}} (y_w | x)}{\pi_{\text{ref}} (y_w | x)} - \beta \log \frac{\pi_{\theta}^{\text{t}} (y_l | x)}{\pi_{\text{ref}} (y_l | x)} \right) \right]
\end{equation}
where $\pi_{\theta}^\text{t}$ represents the actor model $M_t$, and the reference model $\pi_{ref}$ initialized with $M_t$ remains fixed during the training process. This results in a new actor model, $M_{t+1}$, for the next iteration.

\subsubsection{Refiner Training} \label{para: reinfer rft}


Given that the input for the refiner is templated, we use RFT to obtain the new refiner $R_{t+1}$. 
The RFT training data consists of two components: the refinement data and the judgment data for improving the refiner's corresponding capabilities.

\paragraph{Refinement Training Data.} The refinement training data consists of tuples that capture the process of refining incorrect responses. For each incorrect response from the tree-search based refinement step, we collect tuples in the form of $(x, y_p, j_p, y_\text{refined})$, where $(x, y_{p}, j_{p})$ represents the parent node of the final correct response in the refinement tree, and $y_\text{refined}$ is the correctly refined response.

\paragraph{Judgment Training Data.} The judgment training data is derived both from the negative data collection and nodes of the tree-search process. This dataset consists of tuples $(x, y_i, j_i)$, where $x$ is the prompt, $y_i$ is a response to $x$, and $j_i$ is the judgment consistent with majority voting.

Then, we perform supervised fine-tuning using the constructed training data. For the refinement data $D_\text{refine}^t$ we use the tuples $(x, y_p, j_p, y_\text{refined})$ with input $q=(x, y_p, j_p)$ and target $r=y_\text{refined}$. For the judgment data $D_\text{judge}^t$, we use the tuples $(x, y_i, j_i)$ with input $q=(x, y_i)$ and target $r=j_i$. The supervised fine-tuning loss is given by Eq (\ref{loss: SFT}).
By employing this self-play training process with the tree-search based self-refinement strategy, \model iteratively enhances both the actor and refiner models, aiming for continuous self-improvement in instruction-following tasks.

\section{Experiments}
    \subsection{Experiment Setup}

\paragraph{Backbone Models.}
We have conducted experiments on several popular LLMs:
\begin{itemize}[leftmargin=1.5em,itemsep=0pt,parsep=0.2em,topsep=0.1em,partopsep=0.0em]
    \item \textbf{LLaMA3 Series} \citep{llama3} are the best-performing models of their size, showcasing top-tier instruction-following capabilities among open-source LLMs.
    \item \textbf{GLM-4-9B-Chat} \citep{glm2024chatglm} excels in instruction-following tasks, offering competitive performance under 10B parameters.
    \item \textbf{Mistral-7B-Instruct} \citep{jiang2023mistral} is one of the most popular LLMs and has shown good performance across a wide range of tasks.
\end{itemize}

\paragraph{Settings.}
In this work, we focus on enhancing the instruction-following abilities of LLMs in a self-play fashion.
We utilize SFT to bootstrap models under 10B parameters as actor and refiner models. For the more advanced LLaMA3-70B-Instruct, we directly employ it in both roles.
Following this, we perform a three-iteration self-play training using 10k prompts per iteration from our generated dataset. In each iteration, we apply DPO for the actor and RFT for the refiner. 
We refer to the trained LLaMA3-8B-Instruct as \model-8B, LLaMA3-70B-Instruct as \model-70B, GLM-4-9B-Chat as \model-9B, and Mistral-7B-Instruct as \model-7B. 
More implementation details can be found in Appendix \ref{appendix: implement}.
Description of baseline methods is provided in Appendix \ref{appendix: baseline}.


\subsection{Evaluation Benchmarks}
As both the actor and refiner continually evolve within our framework, it's crucial to comprehensively evaluate both of their capabilities.

\paragraph{Actor's Instruction-following Capability.} 
To assess the actor's ability to follow instructions, we rely on two widely-used benchmarks: IFEval \citep{zhou2023instruction} and FollowBench \citep{jiang2023followbench}.
IFEval offers 541 verifiable instructions specifically designed for code-based evaluation. These instructions cover 25 verifiable types, including tasks like \textit{Keyword Frequency} and \textit{Number of Words}.
FollowBench, on the other hand, encompasses five categories of more subjective constraints: \textit{Content}, \textit{Situation}, \textit{Style}, \textit{Format}, and \textit{Example}. This dataset features 820 meticulously curated instructions across five difficulty levels and utilizes a hybrid assessment approach combining rule-based and LLM-as-judge evaluations.

\paragraph{Refiner's Judgment and Refinement Capability.} 
For assessing the refiner's judgment capability, we turn to LLMBar \citep{zeng2023evaluating}, a dataset designed to measure the assessment ability of LLMs in the context of instruction-following tasks. LLMBar includes 419 instruction-response pairs, categorized into two subsets: \textit{Natural} and \textit{Adversarial}. Originally, the task involves pair-wise comparisons to identify successful and failed responses. We adapted it to a point-wise judgment task, asking the model to determine whether each instruction-following task is successful.

To evaluate the refiner's capability in refinement, we split 200 samples from the $D_\text{RSFT}$ to create a test set, and we employ both GPT-4o and \model-8B-RFT-iter3, the refiner after three rounds of training, as judges to evaluate whether the refined responses are accurately following the instructions.

\begin{table}[t]
\renewcommand\arraystretch{1.1}
\centering
\caption{Main results of iteratively trained LLMs on instruction-following benchmarks (Cf. Table \ref{tab:policy_mistral} for full results). P stands for prompt level, and I represents instruction level. L and S denote loose and strict evaluations, respectively. Avg. indicates average results and Lv means level. Results using inference-time tree search are highlighted in \colorbox{backgreen}{green}. The highest results for each backbone model is \textbf{bolded}. Scores marked with $^\dagger$ are sourced directly from the original paper.}
\vspace{1mm}
\resizebox{\linewidth}{!}{
\begin{tabular}{l|ccccc|cccccc}
\toprule
 & \multicolumn{5}{c|}{\textbf{IFEval}} & \multicolumn{6}{c}{\textbf{FollowBench (SSR)}} \\ \cmidrule(l){2-12} 
\multirow{-2}{*}{\textbf{Model}} & \textbf{P (L)} & \textbf{I (L)} & \textbf{P (S)} & \textbf{I (S)} & \textbf{Avg.} & \textbf{Lv-1} & \textbf{Lv-2} & \textbf{Lv-3} & \textbf{Lv-4} & \textbf{Lv-5} & \textbf{Avg.} \\ \midrule
\multicolumn{12}{c}{\textbf{\textit{LLaMA3-8B Models}}} \\ \midrule
LLaMA3-8B-Instruct & 77.6 & 84.5 & 70.6 & 78.9 & 77.9 & 69.4 & 62.2 & 63.1 & 61.9 & 60.9 & 63.5 \\
AutoIF-8B$^\dagger$ & 43.1 & 56.0 & 28.8 & 42.2 & 42.5 & 54.6 & 52.1 & 50.0 & 49.0 & 43.7 & 49.9 \\
SELF & 78.2 & 84.5 & 76.0 & 82.9 & 80.4 & 68.3 & 65.7 & 65.2 & 62.2 & 62.4 & 64.8 \\
Humpback & 72.5 & 80.2 & 70.1 & 78.1 & 75.2 & 66.8 & 66.1 & 67.2 & 60.2 & 62.6 & 64.6 \\
Self-Rewarding & 77.3 & 84.2 & 74.1 & 81.7 & 79.3 & 72.8 & 66.6 & 66.8 & \textbf{64.9} & 64.1 & 67.0 \\
Meta-Rewarding & 77.8 & 84.1 & 75.4 & 82.3 & 79.9 & 73.9 & 71.9 & 66.0 & 62.3 & 62.6 & 67.3 \\ 
\midrule
\model-8B-SFT & 75.4 & 82.5 & 73.4 & 80.6 & 78.0 & 73.9 & 67.4 & 68.1 & 63.1 & 61.3 & 66.8 \\
\model-8B-DPO-iter1 & 78.0 & 84.7 & 75.8 & 82.6 & 80.3 & \textbf{75.3} & 67.7 & 67.6 & 64.7 & 62.3 & 67.5 \\
\model-8B-DPO-iter2 & 78.9 & 85.0 & 77.1 & 83.3 & 81.1 & 73.9 & 71.9 & 69.1 & 64.0 & 62.2 & 68.2 \\
\model-8B-DPO-iter3 & \textbf{79.9} & \textbf{85.4} & \textbf{78.0} & \textbf{83.7} & \textbf{81.8} & 73.0 & \textbf{72.3} & \textbf{70.0} & 64.1 & \textbf{64.7} & \textbf{68.8} \\
\cdashline{1-12}
~~w/ tree search & \cellcolor{backgreen}82.4 & \cellcolor{backgreen}87.5 & \cellcolor{backgreen}79.5 & \cellcolor{backgreen}85.3 & \cellcolor{backgreen}83.7 & \cellcolor{backgreen}73.9 & \cellcolor{backgreen}71.7 & \cellcolor{backgreen}70.3 & \cellcolor{backgreen}66.8 & \cellcolor{backgreen}64.1 & \cellcolor{backgreen}69.4 \\ \midrule
\multicolumn{12}{c}{\textbf{\textit{GLM-4-9B Models}}} \\ \midrule
GLM-4-9B-Chat &	71.5 & 79.9 & 68.0 & 77.2 & 74.2 & 80.8 & 75.1 & 67.4 & 64.3 & \textbf{65.4} & 70.6 \\ \midrule
\model-9B-SFT & 71.5 & 80.5 & 68.8 & 78.1 & 74.7 & 79.4 & 70.9 & 68.2 & 65.1 & 63.7 & 69.5 \\
\model-9B-DPO-iter3 & \textbf{77.3} & \textbf{84.1} & \textbf{73.6} & \textbf{81.4} & \textbf{79.1} & \textbf{82.7} & \textbf{76.7} & \textbf{67.9} & \textbf{68.3} & 64.2 & \textbf{72.0} \\ \midrule
\multicolumn{12}{c}{\textbf{\textit{LLaMA3-70B Models}}} \\ \midrule
LLaMA3-70B-Instruct & 83.7 & 88.9 & 77.1 & 83.8 & 83.4 & 77.1 & 72.5 & 69.4 & 68.7 & 66.3 & 70.8 \\
AutoIF-70B$^\dagger$ & \textbf{85.6} &	\textbf{90.4} &	80.2&	86.7&	85.7 & 71.0 & 67.2 & 66.2 & 64.6 & 63.5 & 66.5 \\ \midrule
\model-70B-DPO-iter3 & \textbf{85.6} & 90.2 & \textbf{81.3} & \textbf{87.3} & \textbf{86.1} & \textbf{80.3} & \textbf{75.7} & \textbf{71.4} & \textbf{73.7} & \textbf{70.5} & \textbf{74.3} \\
\bottomrule
\end{tabular}}
\label{tab:policy_llama3}
\vspace{-5mm}
\end{table}

\subsection{Actor Evaluation Results}

\paragraph{\model significantly improves instruction-following ability.} 
As illustrated in Table \ref{tab:policy_llama3}, the iteratively trained LLMs demonstrate substantial improvements in both the IFEval and FollowBench benchmarks. Remarkably, after three training iterations, \model-8B-DPO-iter3 even surpasses GPT-4-Turbo (81.3\% average accuracy) on IFEval. 
Moreover, incorporating the tree-search refinement technique during the inference stage significantly boosts performance. Additionally, the \model showcases excellent scalability with respect to model size, which substantially enhances the instruction-following abilities of the LLaMA3-70B-Instruct model.
 


\paragraph{\model does not damage general abilities.} 
As shown in Appendix \ref{appendix: general}, we assessed each iteration's performance on general benchmarks, including GSM8k \citep{cobbe2021training}, TriviaQA \citep{joshi2017triviaqa}, MMLU \citep{hendrycks2020measuring}, and HumanEval \citep{chen2021evaluating}. The results indicate that \model maintains or even improves general performance, particularly on GSM8k and HumanEval benchmarks, demonstrating that enhanced instruction-following capabilities support overall LLM alignment.

\begin{wrapfigure}{r}{0.5\textwidth}
    \centering
    \includegraphics[width=\linewidth]{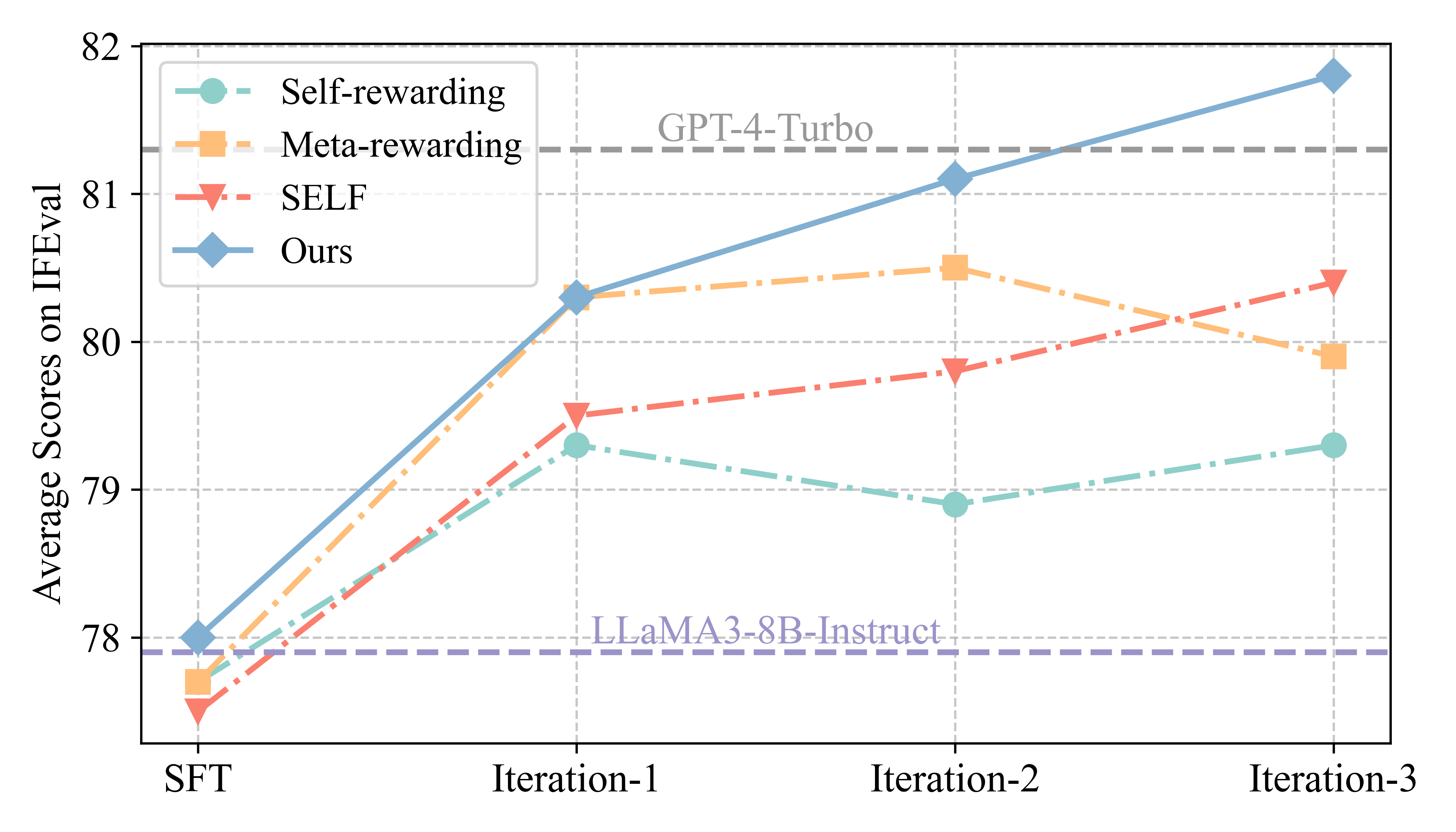}
    \vspace{-8mm}
    \caption{Comparison with baseline methods across iterations (Cf. Figure \ref{fig: mistral baseline} for \model-7B). \model-8B consistently surpasses all baselines.}
    \label{fig: baseline}
    \vspace{-5mm}
\end{wrapfigure}

\paragraph{\model outperforms other baselines significantly.} Figure \ref{fig: baseline} demonstrates the improvements on IFEval with each training iteration. 
In every iteration, \model outperforms other methods.
Notably,

even after three iterations, other methods fail to surpass the performance of \model's first iteration. Generally, our method and SELF outperform self-rewarding and meta-rewarding approaches, underscoring the importance of learning from refinement and excluding the interfering factors in instruction-following tasks. Furthermore, \model's superior performance compared to SELF indicates that contrastive refinement response pairs can highlight key differences, which are difficult to learn using only correct responses.
Additionally, only \model-8B-SFT outperforms the original LLaMA3-8B-Instruct, which suggests that incorporating the judgment SFT or refinement SFT data would reduce performance, likely due to the huge task gap and reduced diversity in the data.

\subsection{Refiner Evaluation Results}

\begin{table}[!t]
\centering
\caption{
Evaluation of judgment capability for iteratively trained LLMs on LLMBar. (Cf. Table \ref{tab:judge_mistral} for Mistral-7B-Instruct results.) Acc. denotes accuracy. The highest scores for each base model are highlighted in \textbf{bold}.}
\vspace{1mm}
\label{tab:judge_llama3}
\renewcommand\arraystretch{1.2}
\resizebox{\linewidth}{!}{
\begin{tabular}{l|cccccccccccccc}
\toprule
\multirow{3}{*}{\textbf{Model}} & \multicolumn{2}{c}{\multirow{2}{*}{\textbf{Natural}}} & \multicolumn{10}{c}{\textbf{Adversarial}} & \multicolumn{2}{c}{\multirow{2}{*}{\textbf{Average}}} \\ \cmidrule(lr){4-13}
 & \multicolumn{2}{c}{} & \multicolumn{2}{c}{\textbf{GPTInst}} & \multicolumn{2}{c}{\textbf{GPTOut}} & \multicolumn{2}{c}{\textbf{Manual}} & \multicolumn{2}{c}{\textbf{Neighbor}} & \multicolumn{2}{c}{\textbf{Average}} & \multicolumn{2}{c}{} \\
 & Acc. & F1 & Acc. & F1 & Acc. & F1 & Acc. & F1 & Acc. & F1 & Acc. & F1 & Acc. & F1 \\ \midrule
GPT-4o-Mini & 74.5 & 70.5 & 69.2 & 61.6 & 60.9 & 51.4 & 59.8 & 51.9 & 72.8 & 66.4 & 65.7 & 57.8 & 67.4 & 60.4 \\ \midrule
\multicolumn{15}{c}{\textbf{\textit{LLaMA3-8B Models}}} \\ \midrule
LLaMA3-8B-Instruct & 60.0 & 51.8 & 55.4 & 46.1 & 47.9 & 39.5 & 51.1 & 36.6 & 54.5 & 45.0 & 52.2 & 41.8 & 53.8 & 43.8 \\ 
SELF & 69.5 & 61.6 & 62.0 & 50.7 & 64.9 & 54.8 & 57.6 & 41.8 & 64.6 & 51.3 & 62.2 & 49.6 & 63.7 & 52.0 \\
Self-Rewarding & \textbf{71.0} & \textbf{66.3} & 70.1 & \textbf{66.7} & 63.8 & 59.5 & 62.0 & 55.7 & 67.5 & 61.7 & 65.9 & 60.9 & 66.9 & 61.9 \\
Meta-Rewarding & 70.5 & \textbf{66.3} & 68.5 & 64.6 & 64.9 & \textbf{60.2} & 64.1 & 58.3 & \textbf{69.0} & \textbf{63.1} & 66.6 & 61.6 & 67.4 & 62.5 \\ \midrule
\model-8B-SFT & 68.5 & 60.9 & 67.9 & 62.4 & 59.6 & 50.0 & 63.0 & 54.1 & 68.3 & 59.3 & 64.7 & 56.5 & 65.5 & 57.3 \\
\model-8B-RFT-iter1 & 68.5 & 63.2 & 66.8 & 60.6 & 63.8 & 55.3 & 62.0 & 53.3 & 66.8 & 59.0 & 64.9 & 57.1 & 65.6 & 58.3 \\
\model-8B-RFT-iter2 & 70.5 & 64.2 & 66.8 & 61.6 & \textbf{66.0} & 60.0 & 65.2 & 57.9 & \textbf{69.0} & 62.4 & 66.8 & 60.5 & 67.5 & 61.2 \\
\model-8B-RFT-iter3 & 70.5 & 65.9 & \textbf{70.7} & \textbf{66.7} & 63.8 & 57.5 & \textbf{68.5} & \textbf{63.3} & 68.3 & 62.2 & \textbf{67.8} & \textbf{62.4} & \textbf{68.3} & \textbf{63.1} \\ \midrule
\multicolumn{15}{c}{\textbf{\textit{GLM-4-9B Models}}} \\ \midrule
GLM-4-9B-Chat & \textbf{74.5} & \textbf{76.5} & 74.5 & \textbf{75.9} & 57.4 & \textbf{62.3} & 53.3 & 56.6 & 69.8 & \textbf{72.0} & 63.7 & \textbf{66.7} & 65.9 & \textbf{68.6} \\ \midrule
\model-9B-SFT & 70.5 & 65.5 & 72.8 & 70.2 & \textbf{59.6} & 55.8 & 64.1 & 53.5 & 71.3 & 67.2 & 66.9 & 61.7 & 67.7 & 62.5 \\
\model-9B-RFT-iter3 & 71.0 & 68.8 & \textbf{75.5} & 74.6 & 58.5 & 55.2 & \textbf{68.5} & \textbf{64.2} & \textbf{68.7} & 65.9 & \textbf{67.8} & 64.9 & \textbf{68.4} & 65.7 \\ \midrule
\multicolumn{15}{c}{\textbf{\textit{LLaMA3-70B Models}}} \\ \midrule
LLaMA3-70B-Instruct & 75.0 & 71.9 & 73.4 & 69.6 & \textbf{69.1} & \textbf{66.7} & 66.3 & \textbf{60.8} & 69.0 & 63.4 & 69.5 & 65.1 & 70.6 & 66.5 \\ \midrule
\model-70B-RFT-iter3 & \textbf{78.0} & \textbf{74.7} & \textbf{78.8} & \textbf{76.9} & 64.9 & 61.2 & \textbf{67.4} & 59.5 & \textbf{72.4} & \textbf{68.1} &\textbf{ 70.9} & \textbf{66.4} & \textbf{72.3} & \textbf{68.1} \\
\bottomrule
\end{tabular}}
\vspace{-3mm}
\end{table}

\paragraph{\model iteratively enhances judgment capability.} Our analysis in Table \ref{tab:judge_llama3} shows that \model iterations notably improve the model's ability to evaluate instruction-following tasks. By iteration three, the refiner \model-8B-RFT-iter3 surpasses GPT-4o-Mini, the model used to construct the judgment SFT dataset. This finding highlights the potential for continuous self-improvement, as the supervised fine-tuning data is not a bottleneck. Interestingly, our refiner greatly outperforms GPT-4o-Mini on adversarial test sets, suggesting that the similar positive and negative examples generated during tree search can make our model more robust against adversarial samples.

\begin{wraptable}{r}{0.4\textwidth}
\centering
\vspace{-5mm}
\caption{Refinement evaluation results. Acc-GPT uses GPT-4o as judge; -\model uses \model-8B-RFT-iter3.}
\vspace{1mm}
\label{tab:refine llama3}
\resizebox{\linewidth}{!}{
\begin{tabular}{l|c|c}
\toprule
\textbf{Model} & \textbf{Acc-GPT}  & \textbf{Acc-\model} \\ \midrule
GPT-4o-Mini        & \textbf{79.0} & 71.0  \\
\midrule
\model-8B-SFT            & 73.5          & 71.0               \\ \midrule
\model-8B-RFT-iter1           & 77.5          & 77.0               \\ \midrule
\model-8B-RFT-iter2           & 74.5          & 76.0               \\ \midrule
\model-8B-RFT-iter3           & \textbf{79.0} & \textbf{90.5}      \\ \bottomrule
\end{tabular}}
\vspace{-3mm}
\end{wraptable}

\paragraph{\model progressively improves refinement capability.} Table \ref{tab:refine llama3} demonstrates continuous improvement in refinement accuracy (success rate) of LLaMA3-8B-Instruct with each training iteration, eventually matching the level of GPT-4o-Mini, the strong LLM for SFT data construction. This further showcases a promising way for self-evolution in instruction-following tasks. However, it also points to a potential issue of self-evaluation bias: when the refiner self-evaluates refinement accuracy, it performs significantly better than when evaluated by GPT-4o.

\subsection{Ablations and Analysis}

\begin{figure}[!t] 
  \begin{minipage}{0.50\textwidth}
    \centering
    \includegraphics[width=\linewidth]{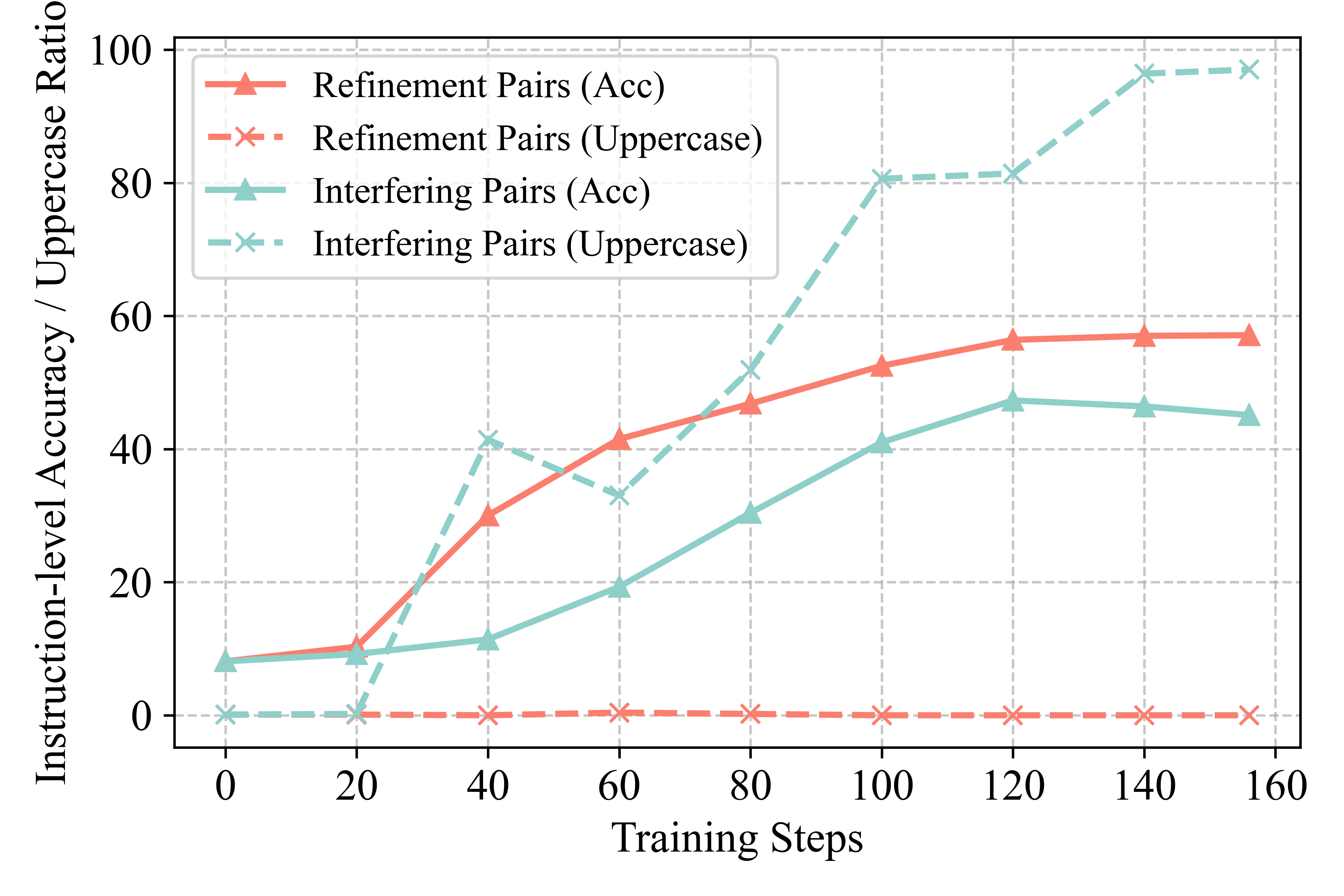}
  \end{minipage}\hfill
  \begin{minipage}{0.49\textwidth}
     \centering
    \includegraphics[width=\linewidth]{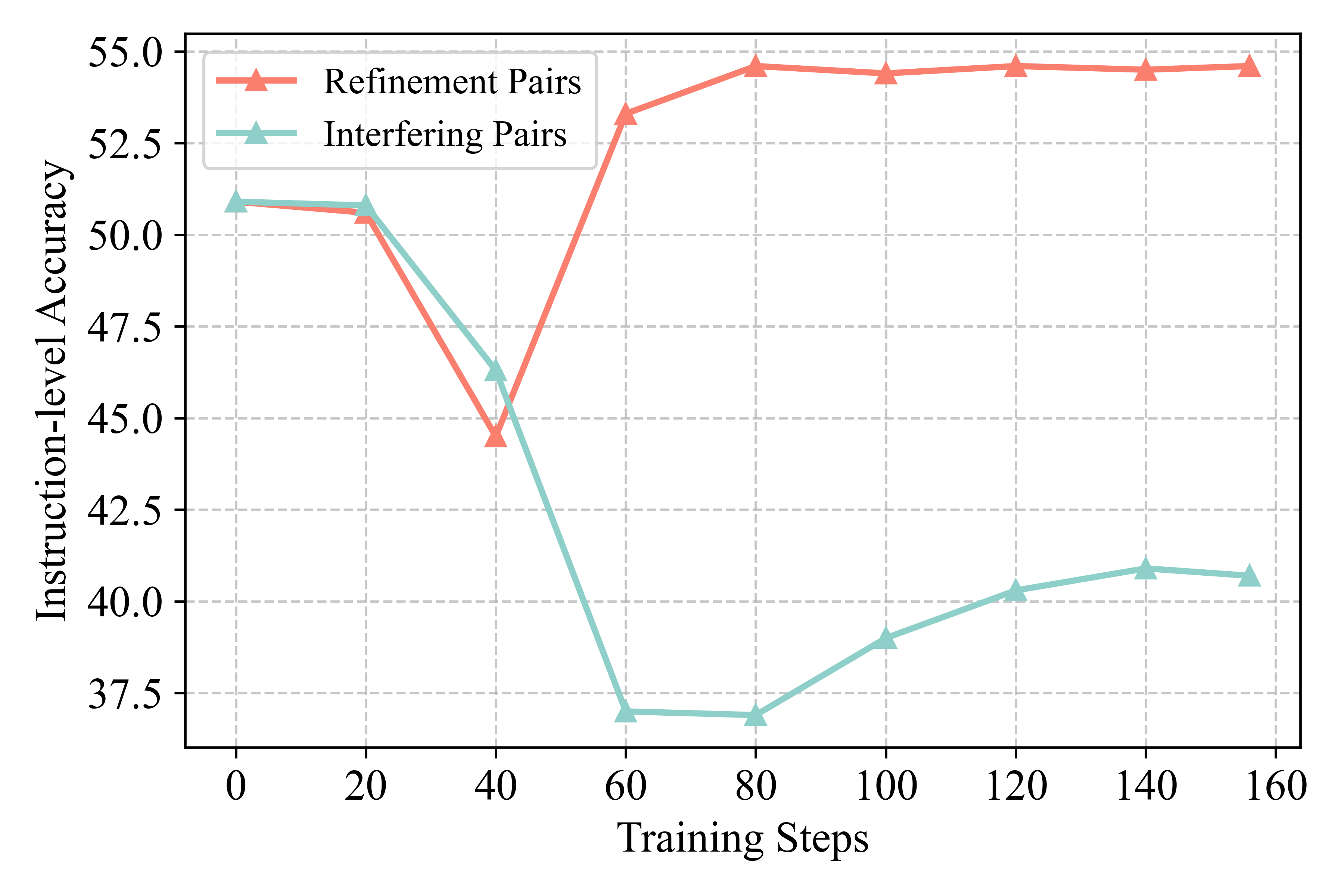}
  \end{minipage}
  \vspace{-10pt}
\caption{Synthetic data experiment results: \textit{Character Sequence Generation} (left) and \textit{Start/End Story Generation} (right).
For \textit{Character Sequence Generation}, interfering pairs show rapid learning of the uppercase ratio (interfering factor) but perform worse than refinement pairs. In the \textit{Start/End Story Generation} task, refinement pairs outperform interfering pairs, which even underperform the original model at step 0.}
\label{fig: synthetic data exp}
\vspace{-7mm}
\end{figure}

\paragraph{Refinement preference pairs enhance instruction-following capability more effectively.}
To verify that the interfering factors indeed affect preference learning and motivate the need to highlight the key differences, we have conducted a synthetic data experiment featuring two tasks:
\begin{itemize}[leftmargin=1.5em,itemsep=0pt,parsep=0.2em,topsep=0.1em,partopsep=0.0em] \item \textit{Character Sequence Generation}: The model needs to generate a specified number of given letters, with no restrictions on letter case, such as generating 12 letters a. For each prompt, we first construct a negative response in lowercase. In order to introduce disturbing factors, we have the correct response in uppercase for interfering pairs while maintaining refined pairs lowercase correctness. 
\item \textit{Start/End Story Generation}: The model is asked to generate a story that starts with sentence 1 and ends with sentence 2. The negative response lacks either sentence 1 or 2. Interfering pairs have a different story concatenated with these sentences; refined pairs keep the same story intact. 
\end{itemize}
Figure \ref{fig: synthetic data exp} shows that refinement pairs significantly outperform interfering pairs in both tasks, with larger and more effective improvements. Particularly in story generation, diverging stories results in worse accuracy than the original model.
Moreover, in the character generation task, we can clearly observe that the interfering factor (uppercase ratio) is learned quickly. However, the task is not performed as well as the refinement setting, highlighting the necessity of focusing on key differences and excluding possible interfering factors.

Furthermore, the ablation study on actor's performance in Table \ref{tab:ablation actor} further reveals a significant drop when refinement data is omitted. \model's superiority over self-rewarding and meta-rewarding methods in Table \ref{tab:policy_llama3} also underscores the importance of using refinement pairs to eliminate interfering factors. Additionally, the string-level similarity of refinement response pairs is 0.90, much higher than 0.85 of the independently sampled response pairs.

\begin{table}[!t]
    \centering

    \begin{minipage}{0.59\textwidth}
    \tiny
    \centering
        \caption{Ablation study on the actor.}
        \label{tab:ablation actor}
    \renewcommand{\arraystretch}{1.1} 
    \setlength{\tabcolsep}{1mm} 
    \resizebox{\linewidth}{!}{
    \begin{tabular}{lccc}
    \toprule
    \multirow{2}{*}{\textbf{Model}} & \multicolumn{2}{c}{\textbf{IFEval}} & \textbf{FollowBench (SSR)} \\
    \cmidrule(lr){2-3} \cmidrule(lr){4-4}
      & \textbf{Prompt(S)} & \textbf{Instruction(S)} & \textbf{Avg.} \\
    \midrule
    \model-8B-DPO-iter3 & 78.0 & 83.7 & 68.8\\
    \midrule
    \textit{w/o} Tree Search & -2.0 & -0.8 & -1.7 \\
    \textit{w/o} Iterative Training & -0.9 & -0.2 & -2.0 \\
    \textit{w/o} Refinement & -2.6  & -1.6   & -3.1 \\
    \bottomrule
    \end{tabular}}
    
    \end{minipage}
    \hfill
    \begin{minipage}{0.4\textwidth}
    \tiny
    \centering
        \caption{Ablation study on the refiner.}
        \label{tab:ablation refiner}
    \renewcommand{\arraystretch}{1.3} 
    \setlength{\tabcolsep}{1.0mm} 
    \resizebox{\linewidth}{!}{
    \begin{tabular}{lcccc}
    \toprule
    \multirow{2}{*}{\textbf{Model}} & \multicolumn{2}{c}{\textbf{Natural}} & \multicolumn{2}{c}{\textbf{Adversarial}} \\
    \cmidrule(lr){2-3} \cmidrule(lr){4-5}
      & Acc. & F1 & Acc. & F1 \\
    \midrule
    \model-8B-RFT-iter3 & 70.5 & 65.9 & 67.8 & 62.4\\
    \midrule
    \textit{w/o} Tree Search & -0.5 & -1.2 & -4.3 & -8.2 \\
    \textit{w/o} Iterative Training & -0.5 & -2.5 & -1.7 & -3.5 \\
    \bottomrule
    \end{tabular}}
    \end{minipage}
    \vspace{-5mm}
\end{table}

\paragraph{Each element is crucial in \model.} The primary elements of \model include the tree-search refinement process and iterative training.
We thus conduct ablation studies to assess the significance of these elements.
For the tree-search process, as shown in Table \ref{tab:ablation actor}, excluding tree search significantly reduces the actor's performance. This might be due to a lack of difficult samples that require more iterations to refine and a reduced number of preference pairs. Table \ref{tab: refine decoding} illustrates that tree search greatly outperforms greedy decoding in response refinement and surpasses other methods, such as best-of-N refinement or simple iterative refinement. Furthermore, tree search is essential for improving judgment capability, especially against adversarial inputs, as indicated in Table \ref{tab:ablation refiner}. Similar responses with opposite labels generated during the tree-search process can enhance robustness against challenging scenarios. Moreover, the results presented in Tables \ref{tab:ablation actor} and \ref{tab:ablation refiner} underscore the importance of iterative training for both the actor and the refiner. This iterative training process ensures mutual improvement, which is crucial for the overall effectiveness of our framework.

\begin{wrapfigure}{r}{0.5\textwidth}
    \vspace{-7mm}
    \centering
    \includegraphics[width=\linewidth]{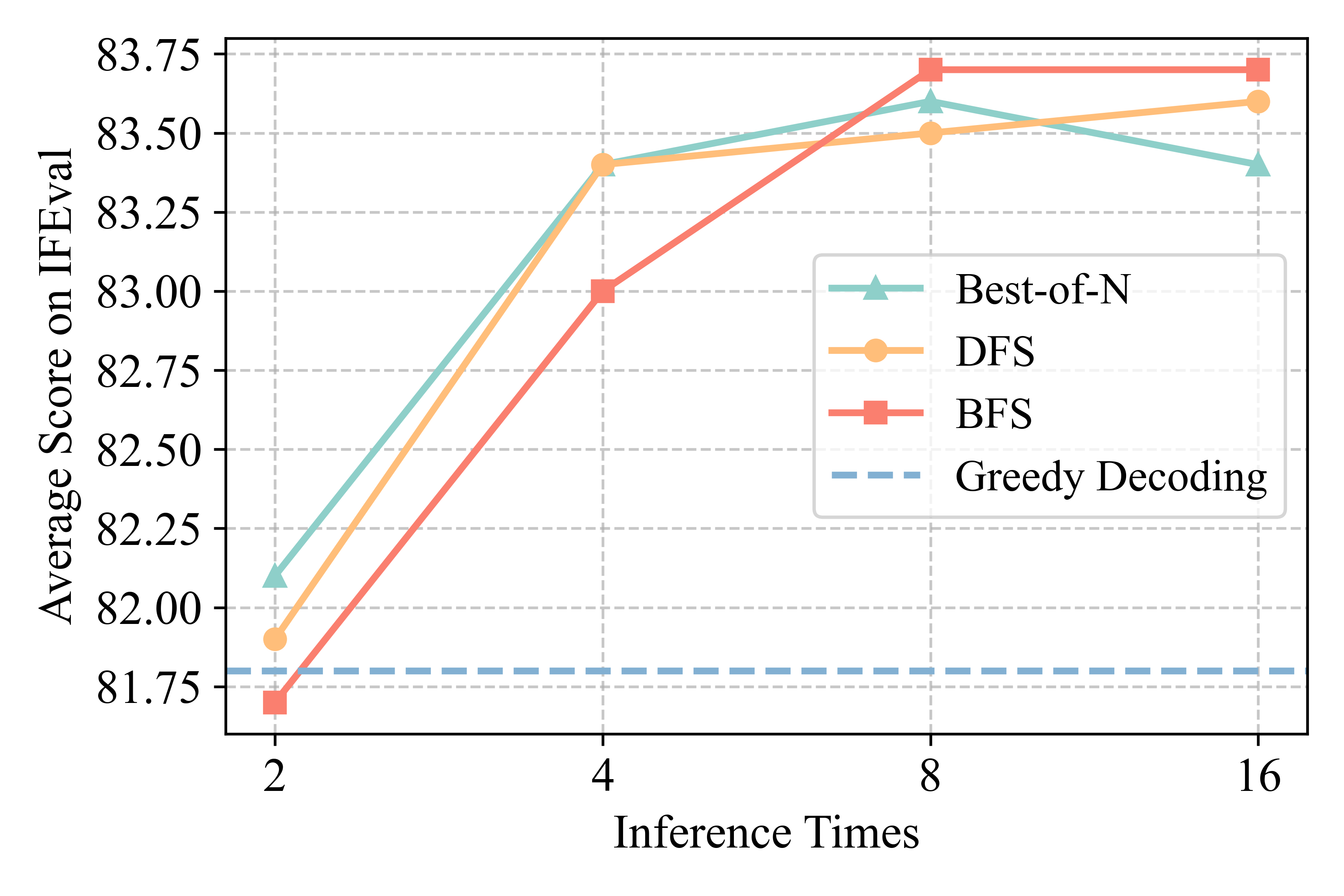}
    \vspace{-10mm}
    \caption{Comparison of decoding strategies.}
    \label{fig: decoding}
    \vspace{-7mm}
\end{wrapfigure}

\paragraph{Scaling test-time compute significantly boosts model performance.} 
Inspired by the recent developments in test-time compute scaling \citep{snell2024scaling}, we investigate various decoding strategies during inference on \model-8B-DPO-iter3.
Figure \ref{fig: decoding} shows that increasing inference times remarkably enhances model performance, outperforming the results of greedy decoding. Notably, while tree search refinement's performance growth is slower, it ultimately achieves superior results compared to best-of-N generation. 
This indicates that refinement is more powerful than generation and could be better suited for scaling test-time compute in the instruction-following task. 










\section{Related Work}
    \subsection{Instruction Following}


Instruction-following is a fundamental capability of LLMs and is central to LLM alignment \citep{ouyang2022training, cheng2023black, lou2024large}. Many studies have evaluated instruction-following capabilities from various perspectives \citep{alpaca_eval, zheng2023judging, zeng2023evaluating, liu2023alignbench, xia2024fofo}. With the expanding application of LLMs, the tasks they are expected to perform become more intricate \citep{liu2023agentbench}, often involving composite instructions with numerous constraints. Consequently, several benchmarks have been developed to test LLMs' ability to follow these complex instructions \citep{zhou2023instruction, jiang2023followbench, qin2024infobench, wen2024benchmarking}. Additionally, multiple studies have focused on enhancing LLMs' instruction-following capabilities \citep{lou2023muffin, zhou2024lima, sun2024conifer}. 
One crucial aspect of the instruction-following task is that subtle differences in responses can significantly impact their correctness \citep{zhou2023instruction}. 
Considering this, we introduce \model framework to construct preference pairs that reduce extraneous elements to highlight these subtle variations for effective improvements.

\subsection{Autonomous LLM Alignment}


Given the high cost of manually collecting alignment data, many studies focus on exploring autonomous LLM alignment methods \citep{cao2024towards}. 
One common strategy involves using data distilled from advanced models to improve less powerful ones \citep{peng2023instruction, xu2023wizardlm, cheng2024autodetect}. 
Alternatively, as the LLMs become stronger, several studies \citep{wang2023self, yuan2024self, zhang2024rest} investigate how to self-evolving LLMs' capabilities. Self-Instruct \citep{wang2023self} generates instructions by employing the model's in-context learning ability. Reinforced Self-Training \citep{gulcehre2023reinforced} samples data from an LLM policy and utilizes the dataset to enhance the policy through offline RL algorithms. 
Moreover, recent research has incorporated feedback from diverse sources. SELF \citep{lu2023self} trains LLMs to acquire meta-skills of self-feedback and self-refinement, enabling the models to self-evolve iteratively. AutoIF \citep{dong2024self} introduces the code execution feedback. Self-rewarding \citep{yuan2024self} and Meta-rewarding \citep{wu2024meta} leverage the LLM-as-judge ability to evaluate its own responses, thereby constructing preference pairs. 
However, these methods usually direct sample multiple independent responses from the actor model, which is likely to introduce the interfering factors and thus affect the model's capture of the key differences.
Thus, we propose a new framework that constructs preference pairs by self-refining the model's responses, minimizing extraneous elements, and promoting more effective autonomous improvement.

\section{Conclusion}
    
In this study, we introduce a new self-play framework, \model, designed to improve the instruction-following capabilities of LLMs through training with refinement pairs. 
We reveal that, unlike traditional approaches that rely on sampling multiple independent responses from the model to construct preference pairs, refining preference pairs to minimize extraneous factors and highlight key differences lead to significant improvements in instruction-following tasks.
Remarkably, the LLaMA3-8B-Instruct model, trained iteratively using our framework, outperforms GPT-4-Turbo on IFEval.
With inference time compute scaling, its performance can be further improved.
Moreover, the iterative enhancement of instruction-following, judgment, and refinement abilities brought about by \model underscores a promising path to continuous self-improvement. 

\section{Acknowledgement}
This work was supported by the National Science Foundation for Distinguished Young Scholars (with No. 62125604). This work was also supported by Tsinghua University Initiative Scientific Research Program.
We would also like to thank Zhipu AI for sponsoring GPU computing and API cost consumed in this study.

\bibliography{iclr2025_conference}

\begin{thebibliography}{49}
\providecommand{\natexlab}[1]{#1}
\providecommand{\url}[1]{\texttt{#1}}
\expandafter\ifx\csname urlstyle\endcsname\relax
  \providecommand{\doi}[1]{doi: #1}\else
  \providecommand{\doi}{doi: \begingroup \urlstyle{rm}\Url}\fi

\bibitem[Bai et~al.(2022)Bai, Jones, Ndousse, Askell, Chen, DasSarma, Drain, Fort, Ganguli, Henighan, et~al.]{bai2022training}
Yuntao Bai, Andy Jones, Kamal Ndousse, Amanda Askell, Anna Chen, Nova DasSarma, Dawn Drain, Stanislav Fort, Deep Ganguli, Tom Henighan, et~al.
\newblock Training a helpful and harmless assistant with reinforcement learning from human feedback.
\newblock \emph{arXiv preprint arXiv:2204.05862}, 2022.

\bibitem[Brown et~al.(2020)Brown, Mann, Ryder, Subbiah, Kaplan, Dhariwal, Neelakantan, Shyam, Sastry, Askell, et~al.]{brown2020language}
Tom Brown, Benjamin Mann, Nick Ryder, Melanie Subbiah, Jared~D Kaplan, Prafulla Dhariwal, Arvind Neelakantan, Pranav Shyam, Girish Sastry, Amanda Askell, et~al.
\newblock Language models are few-shot learners.
\newblock \emph{Advances in neural information processing systems}, 33:\penalty0 1877--1901, 2020.

\bibitem[Cao et~al.(2024)Cao, Lu, Lu, Chen, Ren, Xiang, Liu, Lu, He, Han, et~al.]{cao2024towards}
Boxi Cao, Keming Lu, Xinyu Lu, Jiawei Chen, Mengjie Ren, Hao Xiang, Peilin Liu, Yaojie Lu, Ben He, Xianpei Han, et~al.
\newblock Towards scalable automated alignment of llms: A survey.
\newblock \emph{arXiv preprint arXiv:2406.01252}, 2024.

\bibitem[Chen et~al.(2021)Chen, Tworek, Jun, Yuan, Pinto, Kaplan, Edwards, Burda, Joseph, Brockman, et~al.]{chen2021evaluating}
Mark Chen, Jerry Tworek, Heewoo Jun, Qiming Yuan, Henrique Ponde de~Oliveira Pinto, Jared Kaplan, Harri Edwards, Yuri Burda, Nicholas Joseph, Greg Brockman, et~al.
\newblock Evaluating large language models trained on code.
\newblock \emph{arXiv preprint arXiv:2107.03374}, 2021.

\bibitem[Cheng et~al.(2023)Cheng, Liu, Zheng, Ke, Wang, Dong, Tang, and Huang]{cheng2023black}
Jiale Cheng, Xiao Liu, Kehan Zheng, Pei Ke, Hongning Wang, Yuxiao Dong, Jie Tang, and Minlie Huang.
\newblock Black-box prompt optimization: Aligning large language models without model training.
\newblock \emph{arXiv preprint arXiv:2311.04155}, 2023.

\bibitem[Cheng et~al.(2024)Cheng, Lu, Gu, Ke, Liu, Dong, Wang, Tang, and Huang]{cheng2024autodetect}
Jiale Cheng, Yida Lu, Xiaotao Gu, Pei Ke, Xiao Liu, Yuxiao Dong, Hongning Wang, Jie Tang, and Minlie Huang.
\newblock Autodetect: Towards a unified framework for automated weakness detection in large language models.
\newblock \emph{arXiv preprint arXiv:2406.16714}, 2024.

\bibitem[Chowdhery et~al.(2023)Chowdhery, Narang, Devlin, Bosma, Mishra, Roberts, Barham, Chung, Sutton, Gehrmann, et~al.]{chowdhery2023palm}
Aakanksha Chowdhery, Sharan Narang, Jacob Devlin, Maarten Bosma, Gaurav Mishra, Adam Roberts, Paul Barham, Hyung~Won Chung, Charles Sutton, Sebastian Gehrmann, et~al.
\newblock Palm: Scaling language modeling with pathways.
\newblock \emph{Journal of Machine Learning Research}, 24\penalty0 (240):\penalty0 1--113, 2023.

\bibitem[Cobbe et~al.(2021)Cobbe, Kosaraju, Bavarian, Chen, Jun, Kaiser, Plappert, Tworek, Hilton, Nakano, et~al.]{cobbe2021training}
Karl Cobbe, Vineet Kosaraju, Mohammad Bavarian, Mark Chen, Heewoo Jun, Lukasz Kaiser, Matthias Plappert, Jerry Tworek, Jacob Hilton, Reiichiro Nakano, et~al.
\newblock Training verifiers to solve math word problems.
\newblock \emph{arXiv preprint arXiv:2110.14168}, 2021.

\bibitem[Dong et~al.(2024)Dong, Lu, Li, Xia, Yu, Zhou, and Zhou]{dong2024self}
Guanting Dong, Keming Lu, Chengpeng Li, Tingyu Xia, Bowen Yu, Chang Zhou, and Jingren Zhou.
\newblock Self-play with execution feedback: Improving instruction-following capabilities of large language models.
\newblock \emph{arXiv preprint arXiv:2406.13542}, 2024.

\bibitem[GLM et~al.(2024)GLM, :, Zeng, Xu, Wang, Zhang, Yin, Rojas, Feng, Zhao, Lai, Yu, Wang, Sun, Zhang, Cheng, Gui, Tang, Zhang, Li, Zhao, Wu, Zhong, Liu, Huang, Zhang, Zheng, Lu, Duan, Zhang, Cao, Yang, Tam, Zhao, Liu, Xia, Zhang, Gu, Lv, Liu, Liu, Yang, Song, Zhang, An, Xu, Niu, Yang, Li, Bai, Dong, Qi, Wang, Yang, Du, Hou, and Wang]{glm2024chatglm}
Team GLM, :, Aohan Zeng, Bin Xu, Bowen Wang, Chenhui Zhang, Da~Yin, Diego Rojas, Guanyu Feng, Hanlin Zhao, Hanyu Lai, Hao Yu, Hongning Wang, Jiadai Sun, Jiajie Zhang, Jiale Cheng, Jiayi Gui, Jie Tang, Jing Zhang, Juanzi Li, Lei Zhao, Lindong Wu, Lucen Zhong, Mingdao Liu, Minlie Huang, Peng Zhang, Qinkai Zheng, Rui Lu, Shuaiqi Duan, Shudan Zhang, Shulin Cao, Shuxun Yang, Weng~Lam Tam, Wenyi Zhao, Xiao Liu, Xiao Xia, Xiaohan Zhang, Xiaotao Gu, Xin Lv, Xinghan Liu, Xinyi Liu, Xinyue Yang, Xixuan Song, Xunkai Zhang, Yifan An, Yifan Xu, Yilin Niu, Yuantao Yang, Yueyan Li, Yushi Bai, Yuxiao Dong, Zehan Qi, Zhaoyu Wang, Zhen Yang, Zhengxiao Du, Zhenyu Hou, and Zihan Wang.
\newblock Chatglm: A family of large language models from glm-130b to glm-4 all tools, 2024.

\bibitem[Gulcehre et~al.(2023)Gulcehre, Paine, Srinivasan, Konyushkova, Weerts, Sharma, Siddhant, Ahern, Wang, Gu, et~al.]{gulcehre2023reinforced}
Caglar Gulcehre, Tom~Le Paine, Srivatsan Srinivasan, Ksenia Konyushkova, Lotte Weerts, Abhishek Sharma, Aditya Siddhant, Alex Ahern, Miaosen Wang, Chenjie Gu, et~al.
\newblock Reinforced self-training (rest) for language modeling.
\newblock \emph{arXiv preprint arXiv:2308.08998}, 2023.

\bibitem[Hendrycks et~al.(2020)Hendrycks, Burns, Basart, Zou, Mazeika, Song, and Steinhardt]{hendrycks2020measuring}
Dan Hendrycks, Collin Burns, Steven Basart, Andy Zou, Mantas Mazeika, Dawn Song, and Jacob Steinhardt.
\newblock Measuring massive multitask language understanding.
\newblock \emph{arXiv preprint arXiv:2009.03300}, 2020.

\bibitem[Hou et~al.(2024)Hou, Niu, Du, Zhang, Liu, Zeng, Zheng, Huang, Wang, Tang, et~al.]{hou2024chatglm}
Zhenyu Hou, Yiin Niu, Zhengxiao Du, Xiaohan Zhang, Xiao Liu, Aohan Zeng, Qinkai Zheng, Minlie Huang, Hongning Wang, Jie Tang, et~al.
\newblock Chatglm-rlhf: Practices of aligning large language models with human feedback.
\newblock \emph{arXiv preprint arXiv:2404.00934}, 2024.

\bibitem[Jiang et~al.(2023{\natexlab{a}})Jiang, Sablayrolles, Mensch, Bamford, Chaplot, Casas, Bressand, Lengyel, Lample, Saulnier, et~al.]{jiang2023mistral}
Albert~Q Jiang, Alexandre Sablayrolles, Arthur Mensch, Chris Bamford, Devendra~Singh Chaplot, Diego de~las Casas, Florian Bressand, Gianna Lengyel, Guillaume Lample, Lucile Saulnier, et~al.
\newblock Mistral 7b.
\newblock \emph{arXiv preprint arXiv:2310.06825}, 2023{\natexlab{a}}.

\bibitem[Jiang et~al.(2023{\natexlab{b}})Jiang, Wang, Zeng, Zhong, Li, Mi, Shang, Jiang, Liu, and Wang]{jiang2023followbench}
Yuxin Jiang, Yufei Wang, Xingshan Zeng, Wanjun Zhong, Liangyou Li, Fei Mi, Lifeng Shang, Xin Jiang, Qun Liu, and Wei Wang.
\newblock Followbench: A multi-level fine-grained constraints following benchmark for large language models.
\newblock \emph{arXiv preprint arXiv:2310.20410}, 2023{\natexlab{b}}.

\bibitem[Joshi et~al.(2017)Joshi, Choi, Weld, and Zettlemoyer]{joshi2017triviaqa}
Mandar Joshi, Eunsol Choi, Daniel~S Weld, and Luke Zettlemoyer.
\newblock Triviaqa: A large scale distantly supervised challenge dataset for reading comprehension.
\newblock In \emph{Proceedings of the 55th Annual Meeting of the Association for Computational Linguistics (Volume 1: Long Papers)}, pp.\  1601--1611, 2017.

\bibitem[Li et~al.(2023{\natexlab{a}})Li, Yu, Zhou, Schick, Levy, Zettlemoyer, Weston, and Lewis]{li2023self}
Xian Li, Ping Yu, Chunting Zhou, Timo Schick, Omer Levy, Luke Zettlemoyer, Jason Weston, and Mike Lewis.
\newblock Self-alignment with instruction backtranslation.
\newblock \emph{arXiv preprint arXiv:2308.06259}, 2023{\natexlab{a}}.

\bibitem[Li et~al.(2023{\natexlab{b}})Li, Zhang, Dubois, Taori, Gulrajani, Guestrin, Liang, and Hashimoto]{alpaca_eval}
Xuechen Li, Tianyi Zhang, Yann Dubois, Rohan Taori, Ishaan Gulrajani, Carlos Guestrin, Percy Liang, and Tatsunori~B. Hashimoto.
\newblock Alpacaeval: An automatic evaluator of instruction-following models.
\newblock \url{https://github.com/tatsu-lab/alpaca_eval}, 5 2023{\natexlab{b}}.

\bibitem[Liu et~al.(2024)Liu, Wei, Liu, Si, Zhang, Rao, Zheng, Peng, Yang, Zhou, et~al.]{liu2024best}
Ruibo Liu, Jerry Wei, Fangyu Liu, Chenglei Si, Yanzhe Zhang, Jinmeng Rao, Steven Zheng, Daiyi Peng, Diyi Yang, Denny Zhou, et~al.
\newblock Best practices and lessons learned on synthetic data for language models.
\newblock \emph{arXiv preprint arXiv:2404.07503}, 2024.

\bibitem[Liu et~al.(2023{\natexlab{a}})Liu, Lei, Wang, Huang, Feng, Wen, Cheng, Ke, Xu, Tam, et~al.]{liu2023alignbench}
Xiao Liu, Xuanyu Lei, Shengyuan Wang, Yue Huang, Zhuoer Feng, Bosi Wen, Jiale Cheng, Pei Ke, Yifan Xu, Weng~Lam Tam, et~al.
\newblock Alignbench: Benchmarking chinese alignment of large language models.
\newblock \emph{arXiv preprint arXiv:2311.18743}, 2023{\natexlab{a}}.

\bibitem[Liu et~al.(2023{\natexlab{b}})Liu, Yu, Zhang, Xu, Lei, Lai, Gu, Ding, Men, Yang, et~al.]{liu2023agentbench}
Xiao Liu, Hao Yu, Hanchen Zhang, Yifan Xu, Xuanyu Lei, Hanyu Lai, Yu~Gu, Hangliang Ding, Kaiwen Men, Kejuan Yang, et~al.
\newblock Agentbench: Evaluating llms as agents.
\newblock \emph{arXiv preprint arXiv:2308.03688}, 2023{\natexlab{b}}.

\bibitem[Lou et~al.(2023)Lou, Zhang, Xie, Sun, Ahn, Xu, Su, and Yin]{lou2023muffin}
Renze Lou, Kai Zhang, Jian Xie, Yuxuan Sun, Janice Ahn, Hanzi Xu, Yu~Su, and Wenpeng Yin.
\newblock Muffin: Curating multi-faceted instructions for improving instruction-following.
\newblock \emph{arXiv preprint arXiv:2312.02436}, 2023.

\bibitem[Lou et~al.(2024)Lou, Zhang, and Yin]{lou2024large}
Renze Lou, Kai Zhang, and Wenpeng Yin.
\newblock Large language model instruction following: A survey of progresses and challenges.
\newblock \emph{Computational Linguistics}, pp.\  1--10, 2024.

\bibitem[Lu et~al.(2023)Lu, Zhong, Huang, Wang, Mi, Wang, Wang, Shang, and Liu]{lu2023self}
Jianqiao Lu, Wanjun Zhong, Wenyong Huang, Yufei Wang, Fei Mi, Baojun Wang, Weichao Wang, Lifeng Shang, and Qun Liu.
\newblock Self: Language-driven self-evolution for large language model.
\newblock \emph{arXiv preprint arXiv:2310.00533}, 2023.

\bibitem[MetaAI(2024)]{llama3}
MetaAI.
\newblock Introducing meta llama 3: The most capable openly available llm to date, 2024.
\newblock URL \url{https://ai.meta.com/blog/meta-llama-3}.

\bibitem[Ouyang et~al.(2022)Ouyang, Wu, Jiang, Almeida, Wainwright, Mishkin, Zhang, Agarwal, Slama, Ray, et~al.]{ouyang2022training}
Long Ouyang, Jeffrey Wu, Xu~Jiang, Diogo Almeida, Carroll Wainwright, Pamela Mishkin, Chong Zhang, Sandhini Agarwal, Katarina Slama, Alex Ray, et~al.
\newblock Training language models to follow instructions with human feedback.
\newblock \emph{Advances in neural information processing systems}, 35:\penalty0 27730--27744, 2022.

\bibitem[Peng et~al.(2023)Peng, Li, He, Galley, and Gao]{peng2023instruction}
Baolin Peng, Chunyuan Li, Pengcheng He, Michel Galley, and Jianfeng Gao.
\newblock Instruction tuning with gpt-4.
\newblock \emph{arXiv preprint arXiv:2304.03277}, 2023.

\bibitem[Qin et~al.(2024)Qin, Song, Hu, Yao, Cho, Wang, Wu, Liu, Liu, and Yu]{qin2024infobench}
Yiwei Qin, Kaiqiang Song, Yebowen Hu, Wenlin Yao, Sangwoo Cho, Xiaoyang Wang, Xuansheng Wu, Fei Liu, Pengfei Liu, and Dong Yu.
\newblock Infobench: Evaluating instruction following ability in large language models.
\newblock \emph{arXiv preprint arXiv:2401.03601}, 2024.

\bibitem[Rafailov et~al.(2024)Rafailov, Sharma, Mitchell, Manning, Ermon, and Finn]{rafailov2024direct}
Rafael Rafailov, Archit Sharma, Eric Mitchell, Christopher~D Manning, Stefano Ermon, and Chelsea Finn.
\newblock Direct preference optimization: Your language model is secretly a reward model.
\newblock \emph{Advances in Neural Information Processing Systems}, 36, 2024.

\bibitem[Ruan et~al.(2023)Ruan, Dong, Wang, Pitis, Zhou, Ba, Dubois, Maddison, and Hashimoto]{ruan2023identifying}
Yangjun Ruan, Honghua Dong, Andrew Wang, Silviu Pitis, Yongchao Zhou, Jimmy Ba, Yann Dubois, Chris~J Maddison, and Tatsunori Hashimoto.
\newblock Identifying the risks of lm agents with an lm-emulated sandbox.
\newblock \emph{arXiv preprint arXiv:2309.15817}, 2023.

\bibitem[Shumailov et~al.(2024)Shumailov, Shumaylov, Zhao, Papernot, Anderson, and Gal]{shumailov2024ai}
Ilia Shumailov, Zakhar Shumaylov, Yiren Zhao, Nicolas Papernot, Ross Anderson, and Yarin Gal.
\newblock Ai models collapse when trained on recursively generated data.
\newblock \emph{Nature}, 631\penalty0 (8022):\penalty0 755--759, 2024.

\bibitem[Snell et~al.(2024)Snell, Lee, Xu, and Kumar]{snell2024scaling}
Charlie Snell, Jaehoon Lee, Kelvin Xu, and Aviral Kumar.
\newblock Scaling llm test-time compute optimally can be more effective than scaling model parameters.
\newblock \emph{arXiv preprint arXiv:2408.03314}, 2024.

\bibitem[Sun et~al.(2024)Sun, Liu, Li, Wang, Dong, Lin, and Huang]{sun2024conifer}
Haoran Sun, Lixin Liu, Junjie Li, Fengyu Wang, Baohua Dong, Ran Lin, and Ruohui Huang.
\newblock Conifer: Improving complex constrained instruction-following ability of large language models.
\newblock \emph{arXiv preprint arXiv:2404.02823}, 2024.

\bibitem[Touvron et~al.(2023)Touvron, Lavril, Izacard, Martinet, Lachaux, Lacroix, Rozi{\`e}re, Goyal, Hambro, Azhar, et~al.]{touvron2023llama}
Hugo Touvron, Thibaut Lavril, Gautier Izacard, Xavier Martinet, Marie-Anne Lachaux, Timoth{\'e}e Lacroix, Baptiste Rozi{\`e}re, Naman Goyal, Eric Hambro, Faisal Azhar, et~al.
\newblock Llama: Open and efficient foundation language models.
\newblock \emph{arXiv preprint arXiv:2302.13971}, 2023.

\bibitem[Wang et~al.(2022)Wang, Wei, Schuurmans, Le, Chi, Narang, Chowdhery, and Zhou]{wang2022self}
Xuezhi Wang, Jason Wei, Dale Schuurmans, Quoc Le, Ed~Chi, Sharan Narang, Aakanksha Chowdhery, and Denny Zhou.
\newblock Self-consistency improves chain of thought reasoning in language models.
\newblock \emph{arXiv preprint arXiv:2203.11171}, 2022.

\bibitem[Wang et~al.(2023)Wang, Kordi, Mishra, Liu, Smith, Khashabi, and Hajishirzi]{wang2023self}
Yizhong Wang, Yeganeh Kordi, Swaroop Mishra, Alisa Liu, Noah~A Smith, Daniel Khashabi, and Hannaneh Hajishirzi.
\newblock Self-instruct: Aligning language models with self-generated instructions.
\newblock In \emph{Proceedings of the 61st Annual Meeting of the Association for Computational Linguistics (Volume 1: Long Papers)}, pp.\  13484--13508, 2023.

\bibitem[Wen et~al.(2024)Wen, Ke, Gu, Wu, Huang, Zhou, Li, Hu, Gao, Xu, et~al.]{wen2024benchmarking}
Bosi Wen, Pei Ke, Xiaotao Gu, Lindong Wu, Hao Huang, Jinfeng Zhou, Wenchuang Li, Binxin Hu, Wendy Gao, Jiaxin Xu, et~al.
\newblock Benchmarking complex instruction-following with multiple constraints composition.
\newblock \emph{arXiv preprint arXiv:2407.03978}, 2024.

\bibitem[Wu et~al.(2024)Wu, Yuan, Golovneva, Xu, Tian, Jiao, Weston, and Sukhbaatar]{wu2024meta}
Tianhao Wu, Weizhe Yuan, Olga Golovneva, Jing Xu, Yuandong Tian, Jiantao Jiao, Jason Weston, and Sainbayar Sukhbaatar.
\newblock Meta-rewarding language models: Self-improving alignment with llm-as-a-meta-judge.
\newblock \emph{arXiv preprint arXiv:2407.19594}, 2024.

\bibitem[Xia et~al.(2024)Xia, Xing, Du, Yang, Feng, Xu, Yin, and Xiong]{xia2024fofo}
Congying Xia, Chen Xing, Jiangshu Du, Xinyi Yang, Yihao Feng, Ran Xu, Wenpeng Yin, and Caiming Xiong.
\newblock Fofo: A benchmark to evaluate llms' format-following capability.
\newblock \emph{arXiv preprint arXiv:2402.18667}, 2024.

\bibitem[Xu et~al.(2023)Xu, Sun, Zheng, Geng, Zhao, Feng, Tao, and Jiang]{xu2023wizardlm}
Can Xu, Qingfeng Sun, Kai Zheng, Xiubo Geng, Pu~Zhao, Jiazhan Feng, Chongyang Tao, and Daxin Jiang.
\newblock Wizardlm: Empowering large language models to follow complex instructions.
\newblock \emph{arXiv preprint arXiv:2304.12244}, 2023.

\bibitem[Yuan et~al.(2024)Yuan, Pang, Cho, Sukhbaatar, Xu, and Weston]{yuan2024self}
Weizhe Yuan, Richard~Yuanzhe Pang, Kyunghyun Cho, Sainbayar Sukhbaatar, Jing Xu, and Jason Weston.
\newblock Self-rewarding language models.
\newblock \emph{arXiv preprint arXiv:2401.10020}, 2024.

\bibitem[Yuan et~al.(2023)Yuan, Yuan, Li, Dong, Lu, Tan, Zhou, and Zhou]{yuan2023scaling}
Zheng Yuan, Hongyi Yuan, Chengpeng Li, Guanting Dong, Keming Lu, Chuanqi Tan, Chang Zhou, and Jingren Zhou.
\newblock Scaling relationship on learning mathematical reasoning with large language models.
\newblock \emph{arXiv preprint arXiv:2308.01825}, 2023.

\bibitem[Zeng et~al.(2022)Zeng, Liu, Du, Wang, Lai, Ding, Yang, Xu, Zheng, Xia, et~al.]{zeng2022glm}
Aohan Zeng, Xiao Liu, Zhengxiao Du, Zihan Wang, Hanyu Lai, Ming Ding, Zhuoyi Yang, Yifan Xu, Wendi Zheng, Xiao Xia, et~al.
\newblock Glm-130b: An open bilingual pre-trained model.
\newblock \emph{arXiv preprint arXiv:2210.02414}, 2022.

\bibitem[Zeng et~al.(2023)Zeng, Yu, Gao, Meng, Goyal, and Chen]{zeng2023evaluating}
Zhiyuan Zeng, Jiatong Yu, Tianyu Gao, Yu~Meng, Tanya Goyal, and Danqi Chen.
\newblock Evaluating large language models at evaluating instruction following.
\newblock \emph{arXiv preprint arXiv:2310.07641}, 2023.

\bibitem[Zhang et~al.(2024)Zhang, Zhoubian, Yue, Dong, and Tang]{zhang2024rest}
Dan Zhang, Sining Zhoubian, Yisong Yue, Yuxiao Dong, and Jie Tang.
\newblock Rest-mcts*: Llm self-training via process reward guided tree search.
\newblock \emph{arXiv preprint arXiv:2406.03816}, 2024.

\bibitem[Zhao et~al.(2024)Zhao, Du, Ju, Wu, and Pan]{zhao2024iidoptimizinginstructionlearning}
Hanyu Zhao, Li~Du, Yiming Ju, Chengwei Wu, and Tengfei Pan.
\newblock Beyond iid: Optimizing instruction learning from the perspective of instruction interaction and dependency.
\newblock 2024.
\newblock URL \url{https://arxiv.org/abs/2409.07045}.

\bibitem[Zheng et~al.(2023)Zheng, Chiang, Sheng, Zhuang, Wu, Zhuang, Lin, Li, Li, Xing, et~al.]{zheng2023judging}
Lianmin Zheng, Wei-Lin Chiang, Ying Sheng, Siyuan Zhuang, Zhanghao Wu, Yonghao Zhuang, Zi~Lin, Zhuohan Li, Dacheng Li, Eric Xing, et~al.
\newblock Judging llm-as-a-judge with mt-bench and chatbot arena.
\newblock \emph{Advances in Neural Information Processing Systems}, 36:\penalty0 46595--46623, 2023.

\bibitem[Zhou et~al.(2024)Zhou, Liu, Xu, Iyer, Sun, Mao, Ma, Efrat, Yu, Yu, et~al.]{zhou2024lima}
Chunting Zhou, Pengfei Liu, Puxin Xu, Srinivasan Iyer, Jiao Sun, Yuning Mao, Xuezhe Ma, Avia Efrat, Ping Yu, Lili Yu, et~al.
\newblock Lima: Less is more for alignment.
\newblock \emph{Advances in Neural Information Processing Systems}, 36, 2024.

\bibitem[Zhou et~al.(2023)Zhou, Lu, Mishra, Brahma, Basu, Luan, Zhou, and Hou]{zhou2023instruction}
Jeffrey Zhou, Tianjian Lu, Swaroop Mishra, Siddhartha Brahma, Sujoy Basu, Yi~Luan, Denny Zhou, and Le~Hou.
\newblock Instruction-following evaluation for large language models.
\newblock \emph{arXiv preprint arXiv:2311.07911}, 2023.

\end{thebibliography}
\bibliographystyle{iclr2025_conference}

\appendix
\section{Dataset Information} \label{appendix: prompt and taxonomy}

\begin{figure}[htbp]
    \centering
    \includegraphics[width=0.8\linewidth]{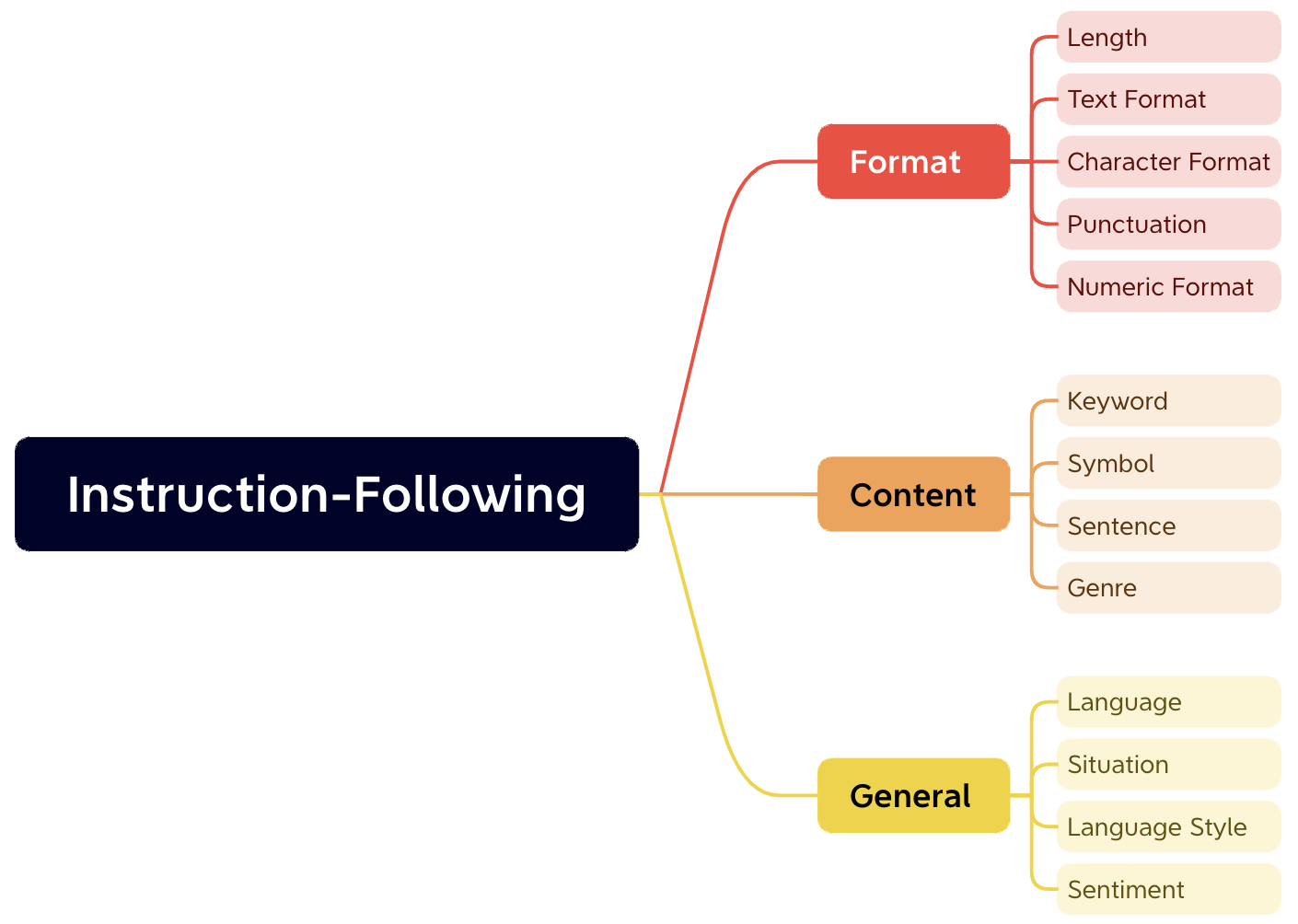}
    \caption{The detailed taxonomy of constraints for prompt evolution.}
    \label{fig: taxonomy}
\end{figure}

\paragraph{Constraint Taxonomy.}
We take the taxonomy from \cite{cheng2024autodetect}, and further refine it to be more comprehensive to ensure the diversity of our prompts. The refined taxonomy is shown in Figure \ref{fig: taxonomy}.

\begin{figure}[htbp]
    \centering
    \includegraphics[width=0.9\linewidth]{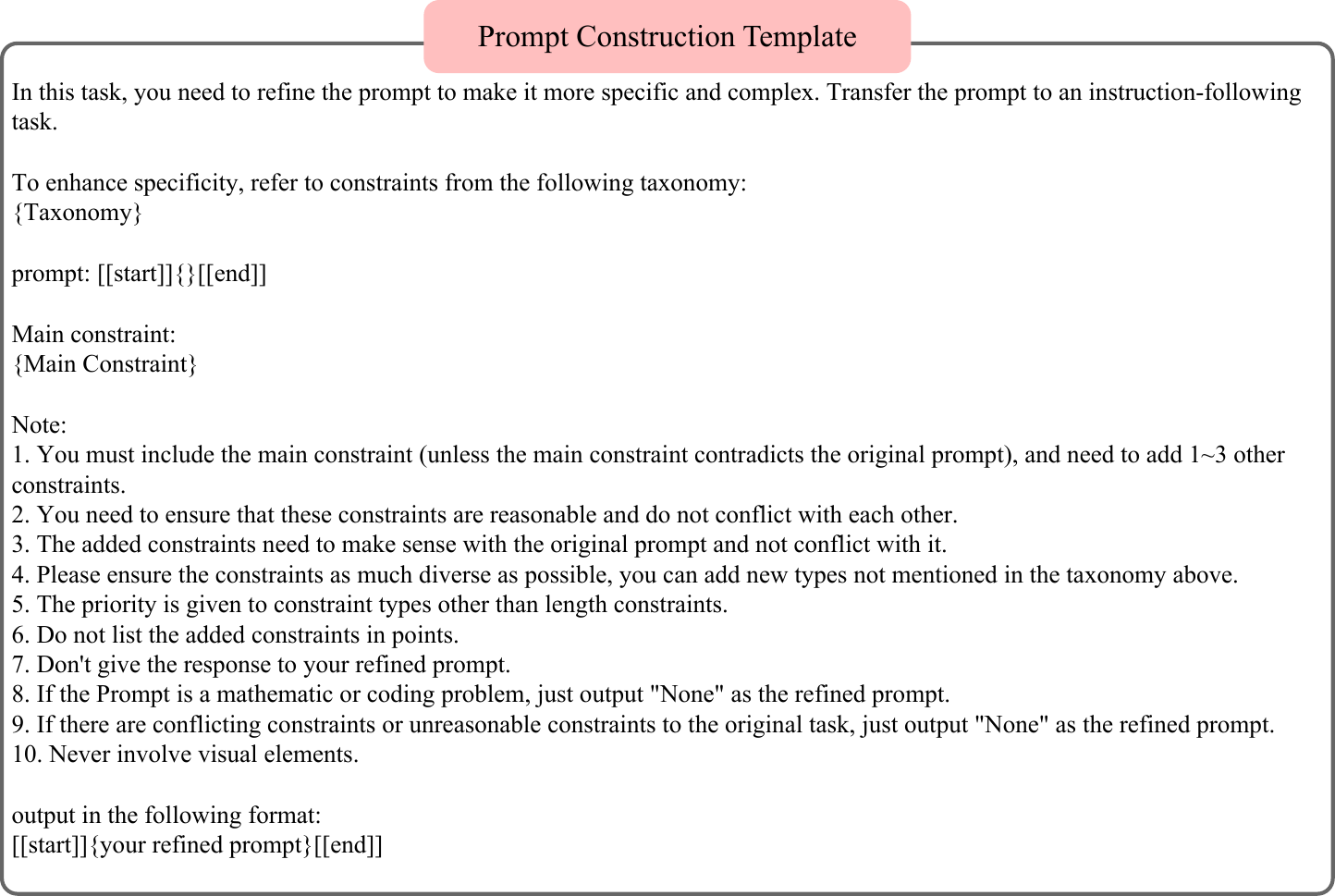}
    \caption{The prompt template applied for prompt evolution.}
    \label{fig: prompt template for evolution}
\end{figure}

\begin{figure}[htbp]
    \centering
    \includegraphics[width=0.9\linewidth]{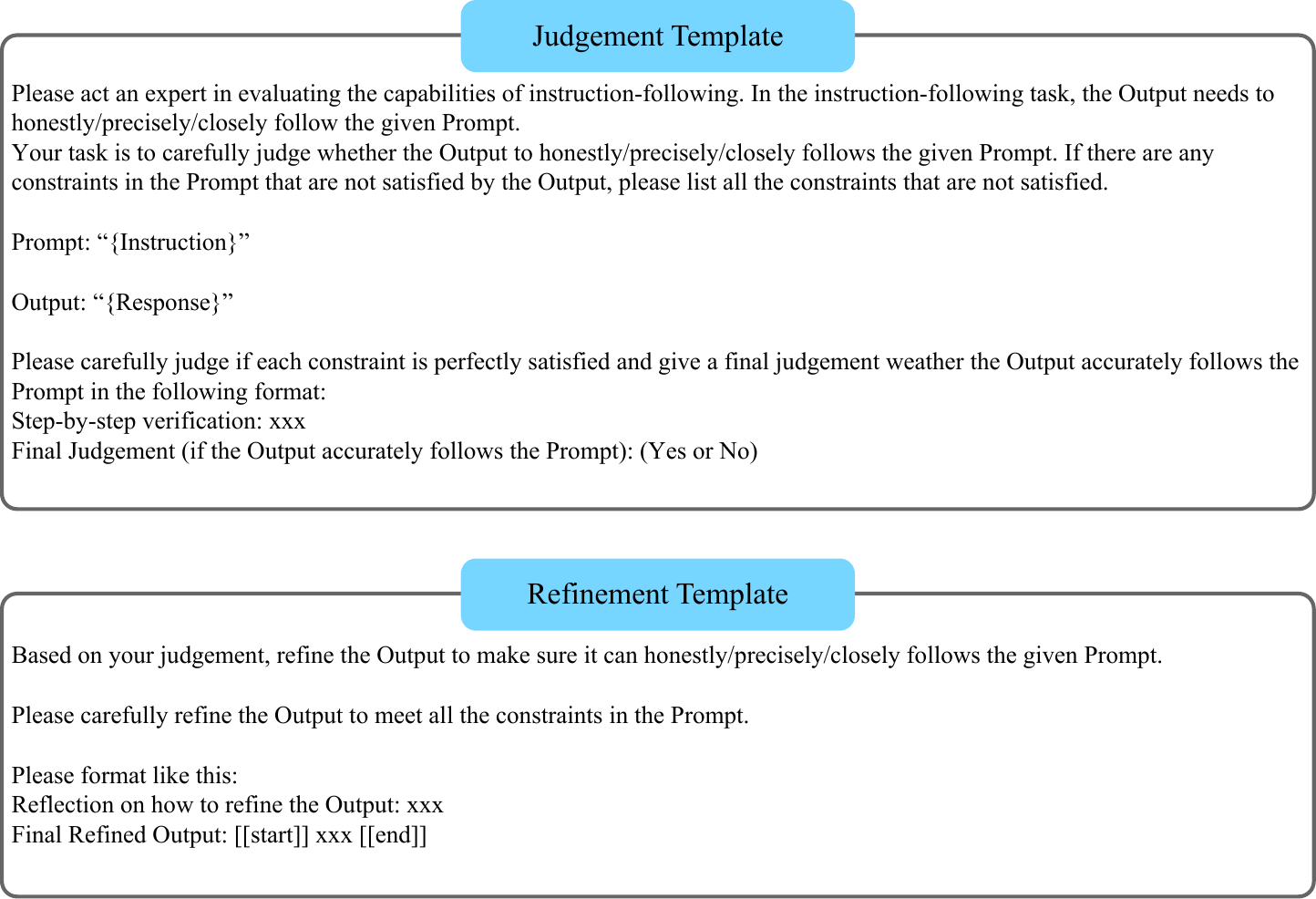}
    \caption{The prompt template applied for the refiner's judgment and refinement.}
    \label{fig: judge prompt template}
\end{figure}

\paragraph{Prompt Template.}
Here, we give the prompt for constructing complex prompts in Figure \ref{fig: prompt template for evolution}.
For the refiner, the prompt template for judgment is provided in Figure \ref{fig: judge prompt template}. As for the refinement task, we form it as a multi-turn task after judgment, with the prompt template provided in Figure \ref{fig: judge prompt template}.

\section{Tree-search Algorithm} \label{appendix: tree search}

We show the detailed process of BFS and DFS refinement in Algorithm \ref{alg:bfs} and Algorithm \ref{alg:dfs}.

\begin{figure}[ht] 
\vspace{-15pt}
  \begin{minipage}{0.50\textwidth}
    \begin{algorithm}[H]
      \caption{BFS-Refinement}
      \label{alg:bfs}
      \begin{algorithmic}
\Require Instruction $x$, Response $y$, Judgment $j$, Refiner $R_{N}$, depth limit $d$, branch limit $b$.

\State $S_0 \gets \{ x, y, j \}$
\For{$t = 1, \cdots, d$}
    \State $S'_t \gets \{ [x, y'] \mid s \in S_{t-1}, y' \in {\color{black}R_N}(s, b) \}$  
    \State $V_t \gets R_N(S'_t)$ \Comment{get judgment}
    \State $S_t \gets \{ [x, y', j'] \mid s \in S'_t, j' \in {\color{black}V_t}(s) \}$
\EndFor \\
\Return $ \arg \max_{s \in S_T} V_T(s)$
\end{algorithmic}
    \end{algorithm}
  \end{minipage}\hfill
  \begin{minipage}{0.49\textwidth}
    \begin{algorithm}[H]
      \caption{DFS-Refinement}
      \label{alg:dfs}
      \begin{algorithmic}
\Require Current state $s$, depth $t$, Refiner $R_N$, depth limit $d$, threshold $v_{\small th}$, branch limit $b$
\If {$t > T$} 
{record output $s = (x, y', j') $} 
\EndIf
\For{$s' \in R_N(s, b)$ } \Comment{refinement}
\If {$R_N(s') < v_{\small th}$}   \Comment{judgment}
\State DFS$(s', t+1)$
\EndIf
\EndFor
\end{algorithmic}
    \end{algorithm}
  \end{minipage}
  \vspace{-9pt}
\end{figure}

\section{Implementation Details} \label{appendix: implement}

The SFT dataset for the actor comprises 8k examples, while the refiner dataset includes approximately 9k examples for judgment training and 3k for refinement training, formatted as a multi-turn task following the first turn's judgment. These two datasets are both constructed with GPT-4o-Mini.
Both the actor and refiner are trained with a learning rate of 2e-6 and a warmup ratio of 0.1, using the AdamW optimizer with $\beta_1=0.9$ and $\beta_2=0.999$. The actor is trained over 5 epochs with a batch size of 64, and the refiner is trained for 3 epochs with the same batch size.
In the data construction process, we set a tree search budget of 15 to strike a balance between performance and efficiency. The average number of expanded tree nodes is around 3.7 in our experiments, which is an acceptable level. Specifically, for LLaMA3-8B-Instruct, the average expanded node numbers are 4.3, 3.7, and 3.4 across different iterations, demonstrating a decreasing trend as the model becomes stronger.
For the actor iterative training, each iteration uses around 5k examples for DPO. To enhance training stability as suggested by \citep{hou2024chatglm}, an additional SFT loss is added to the chosen response with a weight of 0.1. Here, the learning rate is set to 2e-7, $\beta$ to 0.1, with a warmup ratio of 0.1, and training is conducted for 1 epoch with a batch size of 32.
For the refiner, each iteration utilizes about 10k examples, including 4k refinement samples. We ensure the judgment training dataset maintains a balance of positive and negative samples. The training configuration remains the same as for SFT, except the learning rate is set to 1e-6.
All experiments are performed on an 8$\times$80G Nvidia A100 setup.

For our baseline methods, we have maintained uniform settings to ensure fairness. For SELF, we initialize with our constructed datasets, $D_{Actor}$ and $D_{Refiner}$. In the case of self-rewarding and meta-rewarding, we start with $D_{Actor}$ and $D_{JSFT}$. For Humpback, we create the seed dataset by combining about 3k data from the Oasst\footnote{\url{https://huggingface.co/datasets/OpenAssistant/oasst1}} dataset and 5k data from $D_{Actor}$. We also control the number of training samples to be nearly identical for fair comparisons.

\section{Baselines} \label{appendix: baseline}
We compare our method with four popular self-improvement approaches, including: 
\begin{itemize}[leftmargin=1.5em,itemsep=0pt,parsep=0.2em,topsep=0.1em,partopsep=0.0em]
    \item \textbf{AutoIF} \citep{dong2024self} incorporates code feedback and online DPO training to improve instruction-following ability in both distillation and self-evolution settings.
    \item \textbf{SELF} \citep{lu2023self} proposes leveraging language feedback to guide response generation in order to achieve iterative self-improvement.
    \item \textbf{Self-rewarding} \citep{yuan2024self} proposes to combine the reward model and policy model to enhance alignment capabilities simultaneously.
    \item \textbf{Meta-rewarding} \citep{wu2024meta} further introduces a meta-judge to address judgment capability limitations, building on the self-rewarding framework.
    \item \textbf{Humpback} \citep{li2023self} proposes training an instruction generation model to synthesize high-quality data using web resources.
\end{itemize}

\section{Experiment Results} 

\subsection{Instruction-Following Evaluation Results.} 

\begin{figure*}[htbp]
\centering
\resizebox{0.7\textwidth}{!}{
    \includegraphics[width=\linewidth]{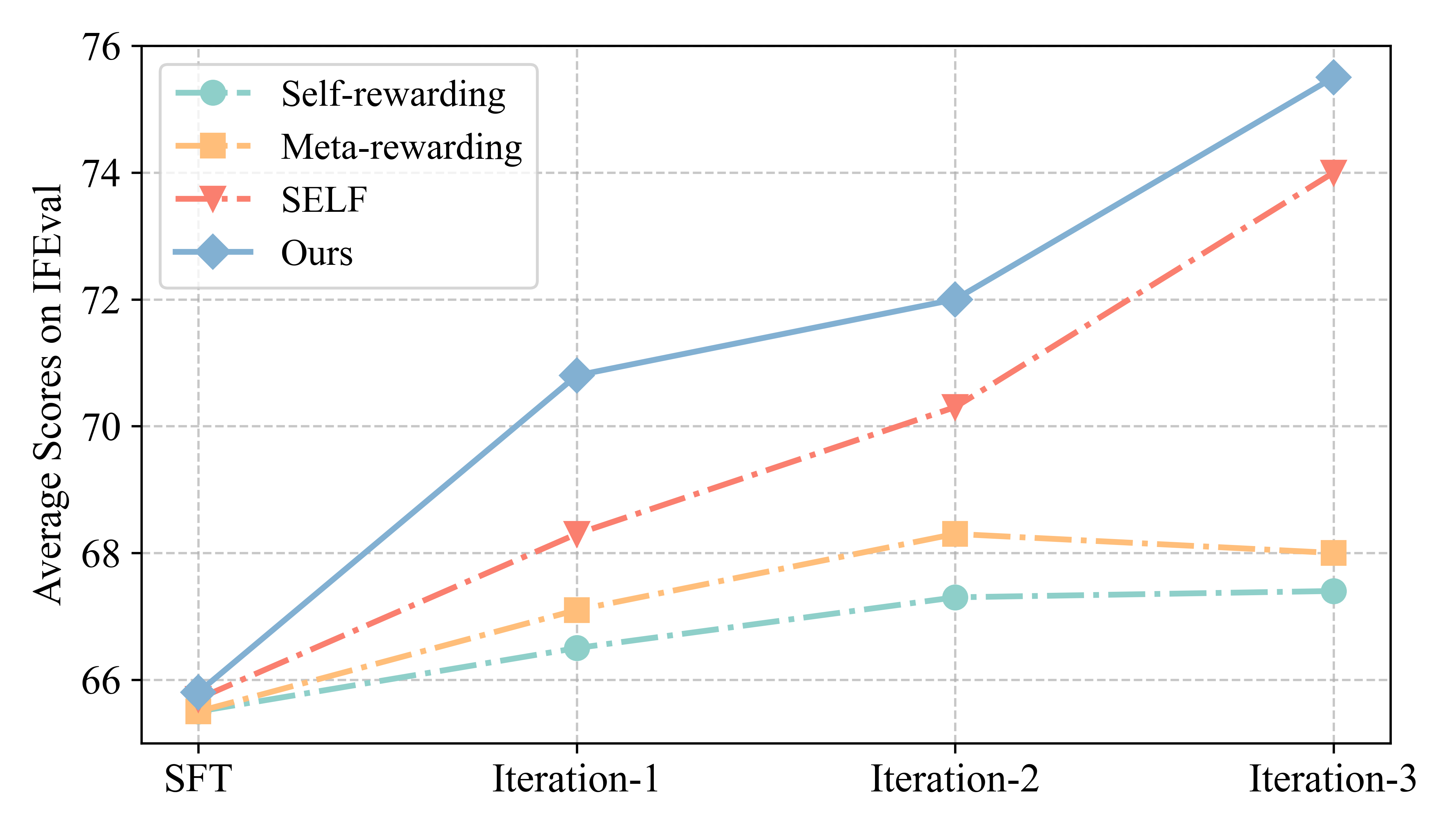}
    }
    \caption{Comparison with baseline methods across iterations. \model-7B consistently surpasses all baselines.}
    \label{fig: mistral baseline}
\end{figure*}

The evaluation results on instruction-following benchmarks are shown in Table \ref{tab:policy_mistral}. Our method outperforms all baselines on these benchmarks and show substantial improvements in each iteration (Figure \ref{fig: mistral baseline}).

\subsection{General Performance Evaluation} \label{appendix: general}
Our analysis in Table \ref{tab: mistral general ability} reveals that \model training not only doesn't harm general performance, but it can also even bring enhancements.

\subsection{Judgment Evaluation Results.} 
As shown in Table \ref{tab:judge_mistral}, the judgment capability improves in each iteration and the accuracy outperforms all baselines.

\subsection{Ablation Study on Judgment Capability.} \label{appendix: exp}
In our experiments, we employ majority voting for iterative improvements for judgment capability. We show the results of the refiner \model-8B-SFT's sampling times and performance on LLMBar in Table \ref{tab:refiner_decoding}. To balance the performance and computation time, we choose majority voting@5.

\subsection{Ablation Study on Refinement Capability.}
Table \ref{tab: refine decoding} shows the results of different decoding strategies for the refinement task on \model-8B. For methods except greedy decoding, we use the same inference budget. We can see that the tree search algorithms largely outperform other methods, verifying the importance of incorporating tree search refinement.

\subsection{Inference-time Scaling Comparison}
\begin{figure}[htbp] 
  \begin{minipage}{0.50\textwidth}
    \centering
    \includegraphics[width=\linewidth]{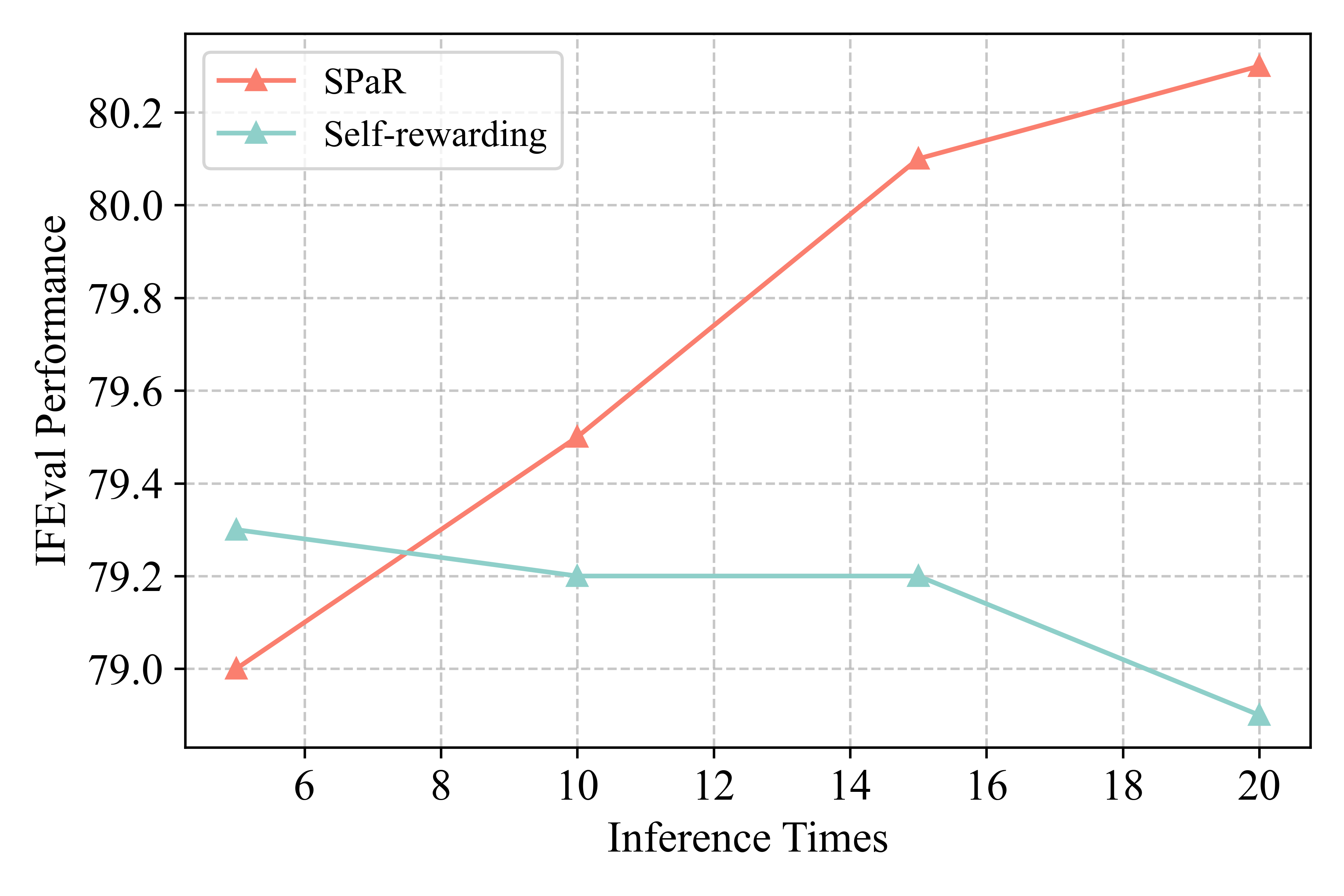}
  \end{minipage}\hfill
  \begin{minipage}{0.49\textwidth}
     \centering
    \includegraphics[width=\linewidth]{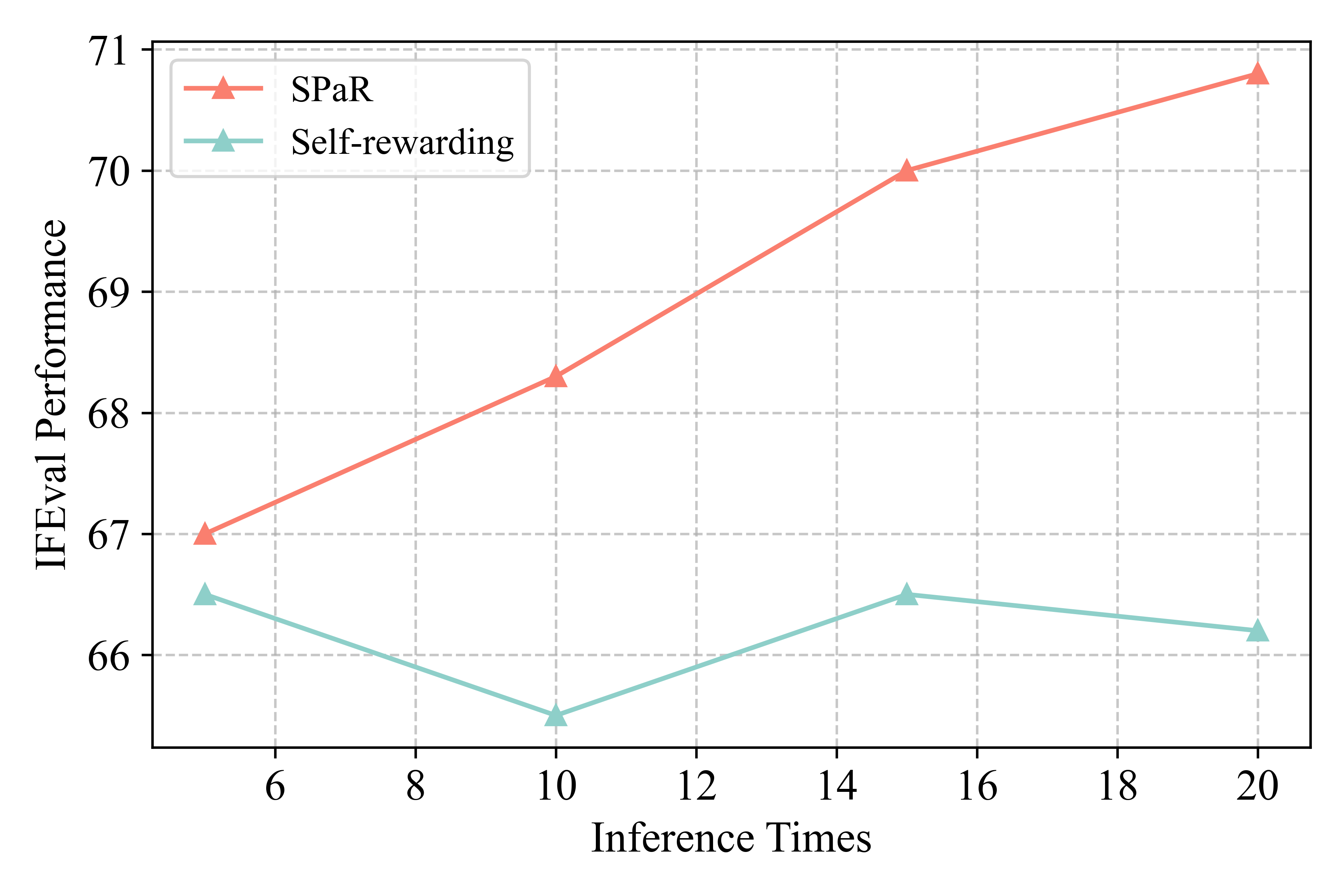}
  \end{minipage}
  \vspace{-10pt}
\caption{Inference-time scaling comparison on IFEval. The left panel showcases results for LLaMA3-8B-Instruct, while the right panel presents findings for Mistral-7B-Instruct.}
\label{fig: inference-time comparison}
\vspace{-4mm}
\end{figure}

Figure \ref{fig: inference-time comparison} presents a comparison between \model and self-rewarding, focusing on their scalability with regard to inference times, measured by the number of response generations in our study. Our analysis includes both the LLaMA3-8B-Instruct and Mistral-7B-Instruct models. The results demonstrate that \model outperforms the self-rewarding method when additional computational resources are allocated for inference time, leading to enhanced performance.

\begin{table}[!t]
\caption{Full results of \model-7B, \model-9B, and \model-70B on instruction-following benchmarks. P stands for prompt level, and I represents instruction level. L and S denote loose and strict evaluations, respectively. Avg. indicates average results and Lv means level. Scores marked with $^\dagger$ are sourced directly from the original paper.}
\vspace{1mm}
\resizebox{\linewidth}{!}{
\begin{tabular}{l|ccccc|cccccc}
\toprule
 & \multicolumn{5}{c|}{\textbf{IFEval}} & \multicolumn{6}{c}{\textbf{FollowBench (SSR)}} \\ \cmidrule(l){2-12} 
\multirow{-2}{*}{\textbf{Model}} & \textbf{P (L)} & \textbf{I (L)} & \textbf{P (S)} & \textbf{I (S)} & \textbf{Avg.} & \textbf{Lv-1} & \textbf{Lv-2} & \textbf{Lv-3} & \textbf{Lv-4} & \textbf{Lv-5} & \textbf{Avg.} \\ \midrule
\multicolumn{12}{c}{\textbf{\textit{Mistral-7B Models}}} \\ \midrule
\rowcolor[HTML]{FFFFFF} 
Mistral-7B-Instruct & 55.1 & 64.9 & 49.9 & 60.2 & 57.5 & 65.1 & 61.6 & 61.6 & 56.8 & 57.2 & 60.4 \\
\rowcolor[HTML]{FFFFFF} 
SELF & 71.3 & 79.7 & 68.0 & 76.9 & 74.0 & 71.5 & 64.2 & 60.8 & 58.0 & 57.0 & 62.3 \\
Humpback & 60.4 & 71.0 & 56.6 & 67.6 & 63.9 & 70.7 & 63.9 & 63.8 & 59.8 & 57.9 & 63.2 \\
Self-Rewarding & 64.3 & 73.5 & 61.0 & 70.7 & 67.4 & 70.8 & 64.8 & 62.3 & 61.9 & \textbf{58.3} & 63.6 \\
\rowcolor[HTML]{FFFFFF} 
Meta-Rewarding & 65.1 & 74.7 & 61.0 & 71.1 & 68.0 & 73.2 & 64.6 & 64.5 & 60.6 & 57.6 & 64.1 \\ \midrule
\rowcolor[HTML]{FFFFFF} 
\rowcolor[HTML]{FFFFFF} 
\model-7B-SFT & 62.7 & 72.3 & 59.3 & 68.7 & 65.8 & 74.4 & 64.3 & 62.5 & 58.2 & 55.0 & 62.9 \\
\rowcolor[HTML]{FFFFFF} 
\model-7B-DPO-iter1 & 68.2 & 76.6 & 64.7 & 73.6 & 70.8 & 73.2 & 64.6 & 63.1 & 60.3 & 56.6 & 63.6 \\
\rowcolor[HTML]{FFFFFF} 
\model-7B-DPO-iter2 & 70.0 & 78.1 & 65.8 & 74.2 & 72.0 & 72.2 & \textbf{65.7} & 61.4 & \textbf{62.4} & 57.5 & 63.8 \\
\rowcolor[HTML]{FFFFFF} 
\model-7B-DPO-iter3 & \textbf{74.1} & \textbf{80.9} & \textbf{69.7} & \textbf{77.1} & \textbf{75.5} & \textbf{74.6} & 63.8 & \textbf{66.1} & 61.0 & 58.0 & \textbf{64.7} \\ \midrule
\multicolumn{12}{c}{\textbf{\textit{GLM-4-9B Models}}} \\ \midrule
GLM-4-9B-Chat &	71.5 & 79.9 & 68.0 & 77.2 & 74.2 & 80.8 & 75.1 & 67.4 & 64.3 & \textbf{65.4} & 70.6 \\ \midrule
\model-9B-SFT & 71.5 & 80.5 & 68.8 & 78.1 & 74.7 & 79.4 & 70.9 & \textbf{68.2} & 65.1 & 63.7 & 69.5 \\
\model-9B-DPO-iter1 & 73.8 & 81.2 & 70.6 & 78.5 & 76.0 & 82.6 & 76.0 & 67.9 & 64.9 & 63.6 & 71.0 \\
\model-9B-DPO-iter2 & 76.7 & 83.3 & 73.2 & 80.9 & 78.5 & 80.4 & 76.6 & 67.4 & \textbf{68.7} & 64.1 & 71.4 \\
\model-9B-DPO-iter3 & \textbf{77.3} & \textbf{84.1} & \textbf{73.6} & \textbf{81.4} & \textbf{79.1} & \textbf{82.7} & \textbf{76.7} & 67.9 & 68.3 & 64.2 & \textbf{72.0} \\ \midrule
\multicolumn{12}{c}{\textbf{\textit{LLaMA3-70B Models}}} \\ \midrule
LLaMA3-70B-Instruct & 83.7 & 88.9 & 77.1 & 83.8 & 83.4 & 77.1 & 72.5 & 69.4 & 68.7 & 66.3 & 70.8 \\
AutoIF-70B$^\dagger$ & \textbf{85.6} &	\textbf{90.4} &	80.2&	86.7&	85.7 & 71.0 & 67.2 & 66.2 & 64.6 & 63.5 & 66.5 \\ \midrule
\model-70B-DPO-iter1 & 84.5 & 89.2 & 80.2 & 85.7 & 84.9 & 77.6 & 74.0 & 70.2 & 70.6 & 66.9 & 71.9 \\
\model-70B-DPO-iter2 & 85.0 & 89.4 & 81.5 & 87.2 & 85.8 & \textbf{80.4} & \textbf{76.4} & 69.9 & \textbf{73.7} & 70.2 & 74.1 \\
\model-70B-DPO-iter3 & \textbf{85.6} & 90.2 & \textbf{81.3} & \textbf{87.3} & \textbf{86.1} & 80.3 & 75.7 & \textbf{71.4} & \textbf{73.7} & \textbf{70.5} & \textbf{74.3} \\
\bottomrule
\end{tabular}}
\label{tab:policy_mistral}
\end{table}
\begin{table}[htbp]
\centering
\caption{Performance on general benchmarks. \model maintains the model's general capabilities.}
\vspace{1mm}
\resizebox{0.8\linewidth}{!}{
\begin{tabular}{l|cccc|l}
\toprule
\textbf{Model}     & \textbf{GSM8k} & \textbf{TriviaQA} & \textbf{MMLU} & \textbf{HumanEval} & \textbf{Average} \\ \midrule
\multicolumn{6}{c}{\textbf{\textit{Mistral-7B Models}}} \\ \midrule
Mistral-7B-Instruct & 42.9	&72.5&	57.9 &	32.9&	51.6         \\ 
\model-7B-SFT                & 56.4&	72.8&	56.7&	44.5&	57.6  {\color{ForestGreen}\textsubscript{\textbf{(+6.0)}}}        \\ 
\model-7B-DPO-iter1               & 55.6&	72.2&	55.3&	46.3&	57.4 {\color{ForestGreen}\textsubscript{\textbf{(+5.8)}}}         \\ 
\model-7B-DPO-iter2               & 54.4	&72.1&	55.8&	45.1&	56.9 {\color{ForestGreen}\textsubscript{\textbf{(+5.3)}}}         \\ 
\model-7B-DPO-iter3               & 58.2	&71.6&	55.1&	46.3&	57.8 {\color{ForestGreen}\textsubscript{\textbf{(+6.2)}}}          \\ \midrule
\multicolumn{6}{c}{\textbf{\textit{LLaMA3-8B Models}}} \\ \midrule
LLaMA3-8B-Instruct & 75.4          & 75.9              & 63.6          & 55.5               & 67.6          \\ 
\model-8B-SFT                & 75.6           & 76.0                & 64.0            & 61.6               & 69.3 {\color{ForestGreen}\textsubscript{\textbf{(+1.7)}}}          \\ 
\model-8B-DPO-iter1               & 78.8          & 75.2              & 63.8          & 60.4               & 69.6 {\color{ForestGreen}\textsubscript{\textbf{(+2.0)}}}        \\ 
\model-8B-DPO-iter2               & 77.0             & 74.9              & 63.1          & 60.4               & 68.9  {\color{ForestGreen}\textsubscript{\textbf{(+1.3)}}}       \\ 
\model-8B-DPO-iter3               & 77.7           & 75.1              & 63.1          & 60.9               & 69.2  {\color{ForestGreen}\textsubscript{\textbf{(+1.6)}}}        \\ \midrule
\multicolumn{6}{c}{\textbf{\textit{GLM-4-9B Models}}} \\ \midrule
GLM-4-9B-Chat & 80.6 & 69.7 & 71.9 & 74.3 & 74.1          \\ 
\model-9B-SFT                & 82.9 & 69.4 & 71.8 & 73.8 & 74.5 {\color{ForestGreen}\textsubscript{\textbf{(+0.4)}}}          \\ 
\model-9B-DPO-iter1               & 82.6 & 68.8 & 71.6 & 75.0 & 74.5 {\color{ForestGreen}\textsubscript{\textbf{(+0.4)}}}        \\ 
\model-9B-DPO-iter2               & 82.8 & 68.9 & 71.8 & 73.8 & 74.3 {\color{ForestGreen}\textsubscript{\textbf{(+0.2)}}}       \\ 
\model-9B-DPO-iter3               & 83.0 & 69.0 & 72.1 & 73.2 & 74.3  {\color{ForestGreen}\textsubscript{\textbf{(+0.2)}}}        \\ \midrule
\multicolumn{6}{c}{\textbf{\textit{LLaMA3-70B Models}}} \\ \midrule
LLaMA3-70B-Instruct & 92.2          & 87.2              & 80.8          & 79.3               & 84.9          \\ 
\model-70B-DPO-iter1                & 92.5 & 90.4 & 81.0 & 79.3 & 85.8 {\color{ForestGreen}\textsubscript{\textbf{(+0.9)}}} \\
\model-70B-DPO-iter2               & 92.9 & 89.5 & 80.4 & 78.7 & 85.4 {\color{ForestGreen}\textsubscript{\textbf{(+0.5)}}} \\
\model-70B-DPO-iter3                & 93.4           & 86.7                & 80.6            & 79.9               & 85.2 {\color{ForestGreen}\textsubscript{\textbf{(+0.3)}}}          \\ 
\bottomrule
\end{tabular}}
\label{tab: mistral general ability} 
\end{table}

\begin{table}[!h]
\caption{Judgment evalution results on LLMBar for \model-7B. Acc. stands for accuracy.}
\resizebox{\linewidth}{!}{
\begin{tabular}{l|cccccccccccccc}
\toprule
\multirow{3}{*}{\textbf{Model}} & \multicolumn{2}{c}{\multirow{2}{*}{\textbf{Natural}}} & \multicolumn{10}{c}{\textbf{Adversarial}} & \multicolumn{2}{c}{\multirow{2}{*}{\textbf{Average}}} \\ \cmidrule(lr){4-13}
 & \multicolumn{2}{c}{} & \multicolumn{2}{c}{\textbf{GPTInst}} & \multicolumn{2}{c}{\textbf{GPTOut}} & \multicolumn{2}{c}{\textbf{Manual}} & \multicolumn{2}{c}{\textbf{Neighbor}} & \multicolumn{2}{c}{\textbf{Average}} & \multicolumn{2}{c}{} \\
 & Acc. & F1 & Acc. & F1 & Acc. & F1 & Acc. & F1 & Acc. & F1 & Acc. & F1 & Acc. & F1 \\ \midrule
Mistral-7B-Instruct & 58.0 & \textbf{69.1} & 57.1 & \textbf{68.8} & 50.0 & \textbf{64.1} & 45.6 & \textbf{61.5} & 47.8 & 62.6 & 50.1 & \textbf{64.3} & 51.7 & \textbf{65.2} \\ 
SELF & 68.0 & 65.2 & 71.2 & 68.7 & 56.4 & 56.8 & 62.0 & 52.6 & 67.5 & 62.3 & 64.3 & 60.1 & 65.0 & 61.1 \\
Self-Rewarding & 68.0 & 64.0 & 69.0 & 63.7 & 59.6 & 53.7 & \textbf{63.0} & 57.5 & \textbf{69.4} & \textbf{64.3} & \textbf{65.3} & 59.8 & 65.8 & 60.6 \\
Meta-Rewarding & 67.5 & 62.4 & 71.7 & 68.7 & 56.4 & 51.8 & \textbf{63.0} & 56.4 & 66.8 & 62.1 & 64.5 & 59.7 & 65.1 & 60.3 \\ \midrule
\model-7B-SFT & 69.5 & 63.9 & 71.7 & 67.5 & 55.3 & 48.8 & 55.4 & 45.3 & \textbf{69.4} & 62.3 & 63.0 & 56.1 & 64.3 & 57.6 \\
\model-7B-RFT-iter1 & 67.0 & 62.1 & 66.3 & 62.7 & 56.4 & 52.9 & 60.9 & 52.6 & 64.2 & 60.7 & 61.9 & 57.2 & 63.0 & 58.2 \\
\model-7B-RFT-iter2 & 68.0 & 64.4 & 68.5 & 64.6 & \textbf{60.6} & 57.5 & 62.0 & 52.1 & 64.2 & 60.0 & 63.8 & 58.5 & 64.7 & 59.7 \\
\model-7B-RFT-iter3 & \textbf{71.0} & 66.7 & \textbf{72.3} & 67.5 & 57.4 & 55.6 & 60.9 & 51.4 & 68.3 & 62.6 & 64.7 & 59.2 & \textbf{66.0} & 60.7 \\ \bottomrule
\end{tabular}}
\label{tab:judge_mistral}
\end{table}
\begin{table}[!t]
\centering

    \caption{Comparison of decoding strategies on LLMBar.}
    \vspace{1mm} 
    \renewcommand{\arraystretch}{1.0} 
    \setlength{\tabcolsep}{1.0mm} 
    \begin{tabular}{lcccc}
    \toprule
    \multirow{2}{*}{\textbf{Method}} & \multicolumn{2}{c}{\textbf{Natural}} & \multicolumn{2}{c}{\textbf{Adversarial}} \\
    \cmidrule(lr){2-3} \cmidrule(lr){4-5}
      & Acc. & F1 & Acc. & F1 \\
    \midrule
    Greedy Decoding & 68.0 & 60.7 & 63.9 & 55.1\\
    \midrule
    Majority Voting@3 & 69.0 & 60.8 & 63.7 & 54.5\\
    \midrule
    Majority Voting@5 & 68.5 & 60.9 & 64.7 & 56.5\\
    \midrule
    Majority Voting@7 & 66.5 & 58.8 & 65.7 & 56.7\\
    \midrule
    Majority Voting@9 & 69.0 & 61.2 & 65.8 & 57.1\\
    \bottomrule
    \end{tabular}
\label{tab:refiner_decoding}
\end{table}

\begin{table}[htbp]
\caption{Comparison of different decoding strategies for refinement task. Acc-GPT stands for the accuracy of using GPT-4o as judge, and Acc-\model for the accuracy of using \model-8B-RFT-iter3 as judge.}
\label{tab: refine decoding}
\vspace{1mm} 
\centering
\begin{tabular}{l|l|l}
\toprule
\textbf{Method}      & \textbf{Acc-GPT} & \textbf{Acc-\model} \\ \midrule
Greedy Decoding      & 69.5             & 65.0                               \\ \midrule
Best of N            & 74.0             & 80.0                               \\ \midrule
Iterative Refinement & 71.0             & 82.0                               \\ \midrule
BFS                  & \textbf{79.0}    & \textbf{90.5}                      \\ \midrule
DFS                  & \textbf{79.0}    & 90.0                               \\ \bottomrule
\end{tabular}
\end{table}

\end{document}